\documentclass[twoside,journal]{IEEEtran}

\usepackage{amsmath,amsfonts,amssymb}
\usepackage{textcomp}
\usepackage{url}
\usepackage{verbatim}
\usepackage{cite}
\usepackage{xcolor}
\usepackage{graphicx}
\usepackage{booktabs}
\usepackage{multirow}
\usepackage{stfloats}
\usepackage{pifont}
\usepackage{algorithm}
\usepackage{algpseudocode}

\usepackage[
    pagebackref=true,
    breaklinks=true,
    colorlinks=true,
    citecolor=blue,
    linkcolor=blue,
    urlcolor=blue,
    hypertexnames=false
]{hyperref}

\usepackage{orcidlink}
\makeatletter
\def\eg{\emph{e.g., }}

\def\ie{\emph{i.e., }}

\makeatother

\newcommand{\cmark}{\ding{51}}
\newcommand{\xmark}{\ding{55}}

\graphicspath{{./image/}{./photos/}}

\newcommand{\rev}[1]{#1}
\newcommand{\slmT}[1]{#1}

\begin{document}
\title{DocPure: Prompt-Free Unified Document Restoration via Degradation-Aware Structure-Guided Wavelet Modulation}

\author{Lingming Su$^{\dagger}$,~Wanglong Lu$^{\dagger}$,~Tao Wang,~Kaihao Zhang,~Nan Zhang,~Liyan An,~Hanli Zhao$^{*}$%
\thanks{Lingming Su, Nan Zhang, Liyan An, and Hanli Zhao are with the College of Computer Science and Artificial Intelligence, Wenzhou University, Wenzhou 325035, China.}
\thanks{Wanglong Lu is with the College of Computer Science and Artificial Intelligence, Wenzhou University, Wenzhou 325035, China, and also with the AI Analytics Team, Nasdaq, St. John's, NL A1A 0L9, Canada.}
\thanks{Tao Wang is with the vivo Mobile Communication Co., Ltd, Shanghai 201210, China.}
\thanks{Kaihao Zhang is with the College of Engineering and Computer Science, The Australian National University, Canberra, ACT, 2601, Australia.}
\thanks{$^{*}$Corresponding author: Hanli Zhao (email: hanlizhao@wzu.edu.cn).}%
\thanks{$^{\dagger}$These authors contributed equally to this work.}
}

\maketitle
\begin{abstract}
 High-quality document images are pivotal for information archiving and downstream automatic processing. However, they are frequently compromised by diverse degradations during uncontrolled acquisition and transmission. While unified document restoration techniques have been proposed to restore images from multiple degradations, they often struggle with {training multiple degradation-specific models, reliance on manual task-specific prompts, or cross-task data pairing.} To address these limitations, we propose DocPure, a  prompt-free unified framework that achieves degradation-aware document restoration. 
 We design a degradation-aware structure auto-encoder \slmT{with degradation-informed routing regularization} to predict clean structural priors from degraded inputs.
\rev{The model is prompt-free at inference, and degradation labels are only used as auxiliary supervision for the routing regularization during training.}
 Furthermore, we introduce a structure-guided wavelet interaction mechanism to bridge frequency-domain features and spatial semantics. Within {the} structure-guided wavelet interaction mechanism, a cross-frequency adaptive modulation utilizes low-frequency sub-bands to modulate high-frequency recovery, ensuring structural consistency. Extensive experiments demonstrate that DocPure achieves {strong performance compared with state-of-the-art methods}  across various tasks, including deblurring, denoising, compression artifact reduction, and deshadowing. The source code will be made publicly available at \href{https://github.com/LingmingSSS/DocPure}
{\texttt{https://github.com/LingmingSSS/DocPure}}.
\end{abstract}


\IEEEpeerreviewmaketitle

\section{Introduction}

Document images frequently suffer from various degradations, such as blur, shadows, and noise.
Unlike natural images, document images are characterized by high-contrast strokes and structured layouts~\cite{appalaraju2021docformer}, making them highly sensitive to such distortions.
These degradations not only reduce visual quality, but also severely degrade downstream OCR performance~\cite{shu2023text,wang2026msinet}. As an active task of image restoration~\cite{zhang2025diff}, document image restoration aims to restore degraded document images to their original legibility.

\begin{figure}[!t]
    \centering
   \includegraphics[width=\columnwidth]{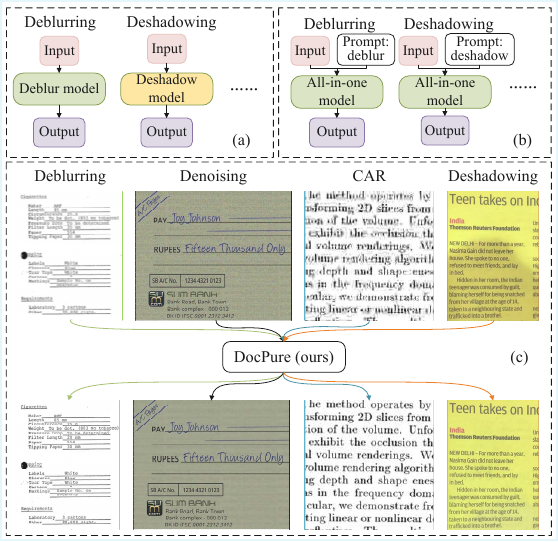}
    \caption{(a) Single-task document restoration or general-task document restoration models require training multiple models for different degradations. (b) Existing unified document restoration models require manual task-specific prompts about the degradation type at the input. (c) For document images afflicted by diverse degradation types, our method consistently outputs clear, high-fidelity images without requiring manual prompts.}
    \label{pic:1}
    \end{figure}

Real-world degradations are diverse and unpredictable \cite{inoue2021learning,liu2024rethinking,jin2025mb,tu2025fourier,tu2026unifying,VSP_2025}.
Given a degraded image, one must typically identify the degradation type first to select the corresponding restoration network.
However, existing methods often suffer from notable limitations. 
As shown in Fig.~\ref{pic:1}~(a) and Fig.~\ref{pic:1}~(b), training separate degradation-specific models~\cite{Li_2023_docshadow,yang2023docdiff} or relying on manual task-specific prompts~\cite{zhang2024docres} is inefficient, while approaches dependent on cross-task data pairing~\cite{wang2024docnlc} are constrained by data scarcity. 
Thus, achieving a unified, prompt-free framework capable of handling diverse degradations remains an open challenge.

\begin{figure}[t]
    \centering
    \includegraphics[width=1\columnwidth]{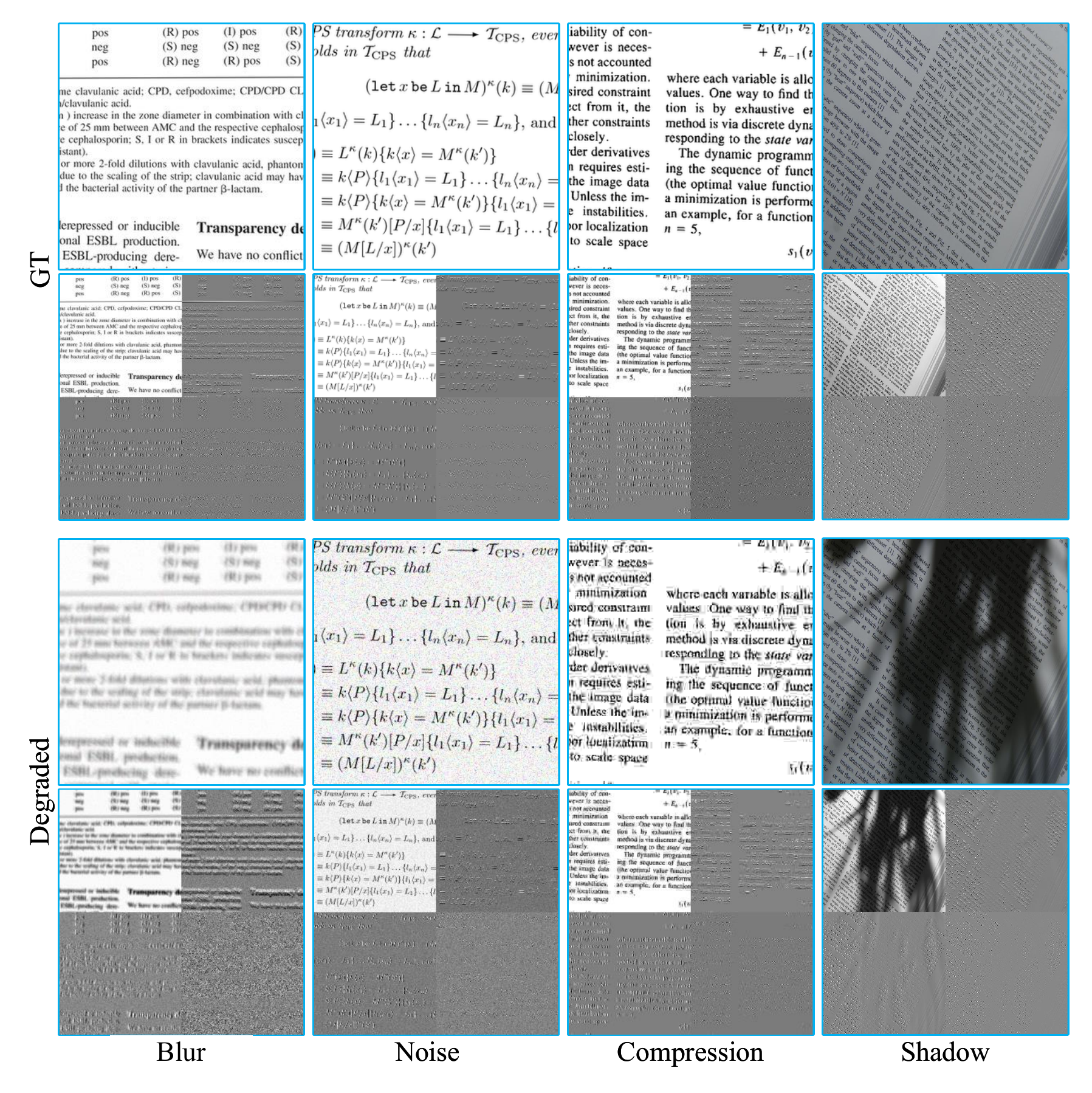}
    \caption{Visualization of document images and their frequency representations under four typical degradation scenarios: blur, noise, compression, and shadow. The first and third rows display the ground truth (GT) and degraded document images in the spatial domain, while the second and fourth rows illustrate their corresponding frequency representations. Each image is decomposed using the Haar wavelet basis~\cite{mallat1989wavelet}, with low-frequency ($L$) and high-frequency ($H$) components obtained in each direction, resulting in four feature sub-bands: $LL$ (top-left), $HL$ (top-right), $LH$ (bottom-left), and $HH$ (bottom-right). These degradations exhibit distinct spectral characteristics in the frequency domain, providing informative cues for decoupling of degradation patterns.}
    \label{fig:wave_energy}
    \end{figure}

To achieve this, \slmT{automatically capturing degradation cues} and restoring text information effectively are two fundamental problems.
As depicted in Fig.~\ref{fig:wave_energy}, we observe that while distinct degradations (\eg blur, noise, compression, shadow) exhibit complex variations in the spatial domain, they manifest unique and distinguishable spectral signatures in the frequency domain. For instance, blur typically {perturbs} high-frequency energy, whereas noise introduces anomalous high-frequency responses. Simultaneously, the legibility of document images hinges on high-contrast strokes and sharp layouts~\cite{appalaraju2021docformer}, which are inherently represented by high-frequency structures.

To this end, we investigate a two-fold strategy to resolve the above two issues.
First, to capture inherent structures without manual prompts, we encode features from degraded images and design a 
\slmT{degradation-informed routing regularization} to {predict clean structural maps}. \rev{This routing mechanism is trained with degradation-label supervision, enabling the model to learn degradation-discriminative representations during training while operating without prompts at inference.}
Second, {using} frequency-domain distributions and predicted structural priors {provides internal cues in place of manual prompts}, thereby enabling the network to discriminate degradation patterns while preserving semantic integrity.

As shown in Fig.~\ref{pic:1}~(c), we propose DocPure, a  prompt-free unified document restoration framework. We first design a degradation-aware structure auto-encoder (DASAE), a U-Net-style multi-encoder routing network for structure extraction. 
DASAE employs a degradation-informed routing mechanism to {produce input-conditioned routing weights for adaptive structure extraction.} 
This \slmT{routing mechanism \rev{facilitates the 
prediction} of clean structural priors}
from inputs affected by varying degradations.
In addition, by leveraging the distinct spectral signatures of different degradations, we devise a mechanism to facilitate the interaction between spatial semantics and frequency-domain representations, thereby ensuring degradation awareness throughout the restoration process. To realize this, we introduce the structure-guided wavelet interaction mechanism. 
First, the mechanism decomposes both the degraded input and the predicted structural map into low- and high-frequency sub-bands. Considering that fine-grained details in document images are predominantly manifested in high frequencies, whereas the low-frequency component of the structure map primarily reflects binary regional intensity, we selectively incorporate the high-frequency components to reinforce fine-grained details.
Subsequently, we propose the cross-frequency adaptive modulation (CFAM) in the \rev{proposed wavelet interaction mechanism} to leverage the structural and degradation priors latent in the low-frequency sub-band, which adaptively modulates fine-grained details within the corresponding high-frequency sub-bands. Throughout the process, the mechanism facilitates the interaction between frequency- and spatial-domain features via a cross-attention mechanism, ensuring holistic degradation awareness and structural consistency in the restoration.

Extensive experiments conducted on 12 diverse datasets across four representative degradation scenarios, including document deblurring, denoising, compression artifact reduction, and deshadowing, validate that DocPure achieves performance \slmT{comparable} to or surpassing state-of-the-art (SOTA) approaches, offering a practical solution for document image restoration.
In summary, the main contributions of this paper are as follows:

\begin{itemize}

\item We introduce DocPure, \rev{a novel unified document image restoration framework that achieves adaptive high-quality restoration} without requiring manual prompts at inference.

\item We propose a degradation-informed routing regularization mechanism to construct a degradation-aware structure auto-encoder \rev{that guides feature routing and facilitates structure extraction}.

\item We introduce a structure-guided wavelet interaction mechanism with cross-frequency adaptive modulation, which leverages spatial-frequency interactions to restore text details while suppressing over-restoration artifacts.

\end{itemize}

\section{Related work}

\slmT{Recent progress in general image restoration has shown that Transformers are effective in modeling complex degradations, such as non-local artifacts \cite{xiao2023random}, restoration uncertainty \cite{xiao2026bayesian}, and illumination changes \cite{xiao2024homoformer}. 
These works suggest the need for degradation-aware modeling, which is also important for document restoration.}
Existing document restoration research primarily follows two paradigms.
Single-task document restoration methods are tailored for specific degradation types, while unified document restoration methods are designed to handle multiple degradations within a general-task or all-in-one architecture.

\begin{figure*}[t]
    \centering
    \includegraphics[width=\textwidth]{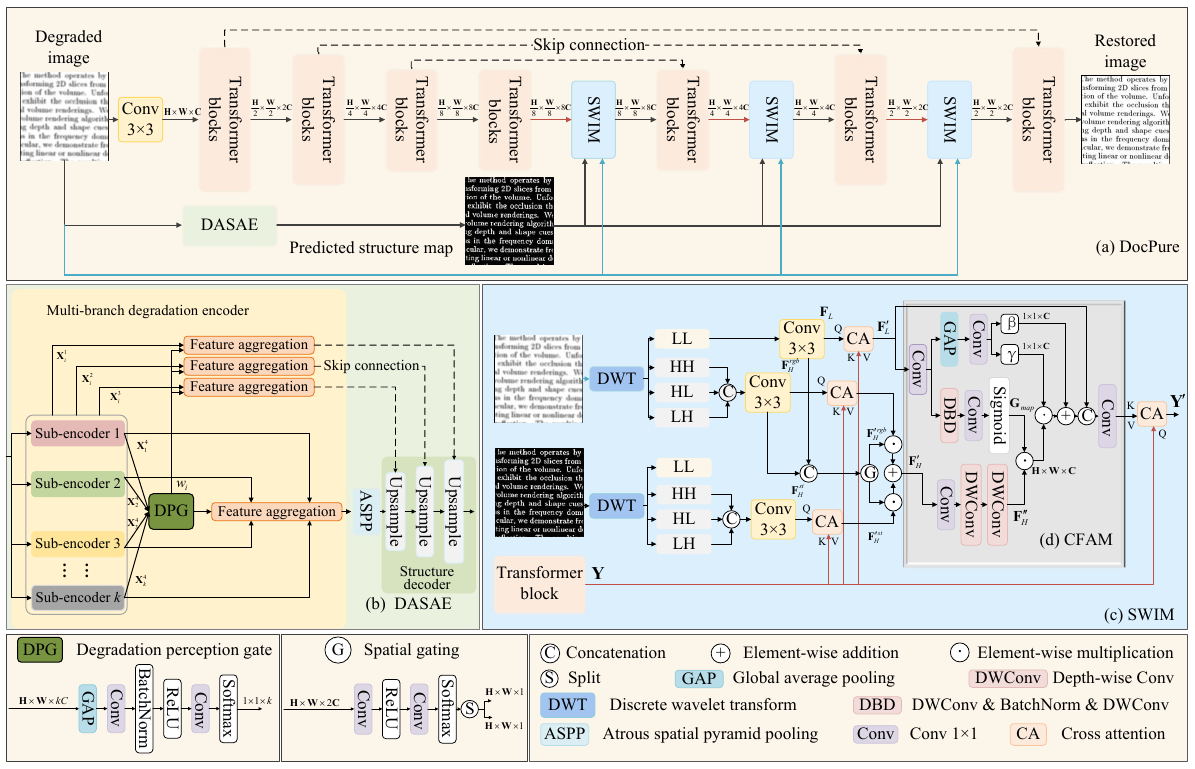}
    \caption{Overview of the proposed DocPure framework. 
    (a) Our method employs a dual-branch architecture. 
    (b) The degradation-aware structure auto-encoder (DASAE) extracts a clear structure map from the degraded input to guide the Transformer-based restoration network. 
    (c) The structure-guided wavelet interaction module (SWIM) facilitates spatial-frequency interactions to enhance degradation awareness. 
    (d) Cross-frequency adaptive modulation (CFAM) reinforces high-frequency details using low-frequency priors, 
    \slmT{helping prevent}
    over-restoration artifacts while ensuring structural sharpness.}
    \label{model1}
\end{figure*}

\textbf{Single-task document restoration.}
For \textbf{deblurring}, research primarily focuses on recovering high-frequency details under unknown degradation conditions. {Blur2sharp~\cite{neji2021blur2sharp}  generates sharp document images without requiring prior knowledge of the blur kernel. DeblurGAN-CNN~\cite{deblureGAN-cnn} improves character legibility by integrating adversarial training with CNN models. DeepDeblur~\cite{mei2019deepdeblur} learns both low-level pixel features and high-level semantic features to address complex degradation mappings.}
In terms of \textbf{denoising}, FFDNet~\cite{zhang2018ffdnet} handles spatially variant Gaussian noise by taking tunable noise maps as input, and may inadvertently over-smooth fine character strokes. {Doc-Attentive-GAN~\cite{neji2024doc} employs an attention-driven GAN to target local reconstruction errors and preserve glyph topology.}
For \textbf{compression artifact reduction}, removing blocking and ringing effects is critical to enhancing the visual quality of text. DMCNN~\cite{zhang2018dmcnn} exploits correlations in both pixel and DCT domains to effectively recover information lost during quantization. TPGSR~\cite{ma2023text} incorporates categorical text priors to guide the restoration process, leveraging semantic guidance to ensure the topological correctness of reconstructed characters.
For \textbf{deshadowing}, \rev{representative methods} incorporate physical priors by introducing frequency-aware modules~\cite{Li_2023_docshadow,UHDRes} or bilinear imaging models~\cite{liu2023shadow_tnnls}. Recently, attention-oriented detail recovery networks~\cite{yang2025adr} and color-aware strategies~\cite{zhang2023RDD} have been proposed to enhance texture fidelity after document restoration. 
To address the bottleneck of the requirement for paired data, studies like FSD~\cite{matsuo2022FSD} explore synthetic data generation schemes based on foreground detection. Although these methods are effective in single-task scenarios, they fail to share knowledge and feature representations across different tasks.  In contrast, our {prompt-free} all-in-one design enables a single prompt-free restoration model \slmT{to handle multiple degradation types observed during training}, without manual prompts or model selection.

\textbf{General-task document restoration.}
Initial research focused on designing robust network topologies capable of adapting to various tasks, though often requiring separate training.
DE-GAN~\cite{souibgui2020gan} adopts a paired data-driven supervised learning strategy to achieve the refined removal of both document background shadows and noise. 
DocDiff~\cite{yang2023docdiff} introduces diffusion models into document image restoration, achieving robust recovery by predicting the residual between the degraded image and the GT distribution. 
TextDoctor~\cite{lu2025textdoctor} employs a patch pyramid to achieve low-cost restoration for high-resolution images. 
Recently, Docstain~\cite{li2025high} extracts degradation features at different scales and fuses them using a dual-attention mechanism to enhance restoration performance. 
However, while effective, these methods typically necessitate training independent parameters for each specific task. Consequently, they lack cross-task feature sharing and joint optimization, failing to achieve true multi-task unification.

\textbf{All-in-one document restoration.}
Several all-in-one frameworks have been proposed to handle multiple degradations within a single trained model.
DocNLC~\cite{wang2024docnlc} is a representative outcome of this research. However, its contrastive learning design necessitates training datasets containing multiple degradations paired with a single GT. This limitation exacerbates the difficulty of data acquisition in real-world applications. 
DocRes~\cite{zhang2024docres} employs a unique prompt generator to produce distinct priors for different degradations, thereby enabling the model to process diverse degradations differentially. Nevertheless, it remains reliant on the manual provision of task-specific prompts to inform the model of the degradation type. 
Uni-DocDiff~\cite{zhao2025unidocdiff} utilizes a unified prior pool to store fixed prior features extracted from input images, aiming to mitigate task interference. However, the model still requires explicit task prompts to select the most appropriate prior from the pool. In contrast, we aim to explore intrinsic features for various degradations to achieve degradation-aware restoration.

\section{Method}  

\subsection{Overview}

We present DocPure, a \rev{unified framework for document image restoration} across diverse degradation scenarios. Unlike prior approaches, DocPure eliminates the need for manual task-specific prompts \slmT{at inference} or {cross-task data pairing}. Instead, it leverages \rev{degradation-informed routing} to achieve unified, cross-task restoration within a single model. DocPure directly reconstructs a clean and legible document image from the degraded image.

{As illustrated in Fig.~\ref{model1}~(a), the overall architecture of DocPure comprises two core networks: a degradation-aware structure auto-encoder (DASAE) and a restoration network. The backbone of the restoration network adopts a classic 4-level U-shaped encoder-decoder architecture, where each level of the encoder and decoder is composed of multiple Transformer blocks~\cite{zamir2022restormer} and our structure-guided wavelet interaction modules (SWIM). Given an input degraded document image $\mathbf{I} \in \mathbb{R}^{H \times W \times 3}$ with height $H$ and width $W$, a $3\times3$ convolutional layer extracts shallow features from $\mathbf{I}$, which are then forwarded to the backbone. In the meanwhile, DASAE  (Fig.~\ref{model1}~(b)) predicts a structure map ${\mathbf{I}_M \in \mathbb{R}^{H \times W \times 1}}$ from $\mathbf{I}$ by identifying diverse degradation patterns. Then, both the degraded image $\mathbf{I}$ and the predicted structure map ${\mathbf{I}_M}$ are fed into the restoration network.  Within SWIM (Fig.~\ref{model1}~(c)), the interaction between wavelet-domain and spatial-domain features is performed using ${\mathbf{I}_M}$ and ${\mathbf{I}}$. It establishes a cross-domain interaction mechanism to explicitly capture degradation characteristics within the spectral domain. As a key module in SWIM, CFAM is designed to balance high-frequency detail enhancement with low-frequency structural constraints. Finally, the network outputs the high-quality restored document image $\hat{\mathbf{I}}$.}

\subsection{Degradation-aware structure auto-encoder}

Document structure is a critical prior for restoration as it preserves text morphology. However, designing specialized extractors for different degradations is inefficient and limits generalization. To address this, we propose DASAE, which automatically extracts a unified structure prior across diverse degradation types.

As illustrated in Fig.~\ref{model1}~(b), our DASAE first employs a multi-branch \slmT{degradation} encoder with a degradation perception gate (DPG) to \slmT{model degradation-conditioned features and generate perception confidence scores.} These scores then guide feature aggregation, dynamically fusing shallow features into the structure decoder to \slmT{enhance degradation-aware structure prediction. We further introduce a degradation-informed routing regularization to encourage the routing distribution to be consistent with the degradation-label prior during training.}

\begin{figure}[t]
    \centering
    \includegraphics[width=\columnwidth]{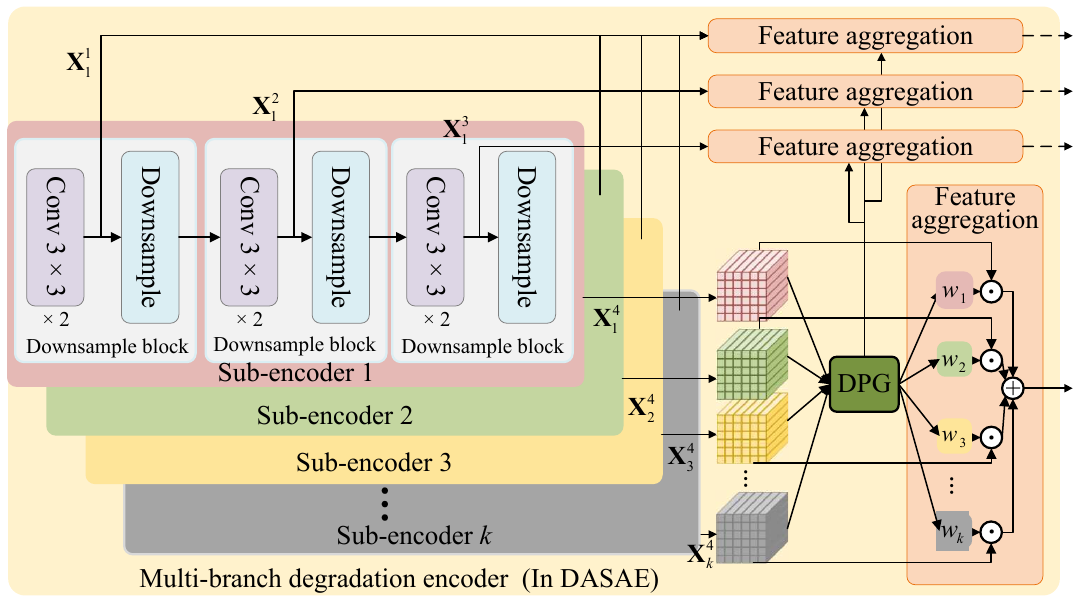}
    \caption{
    Details of the multi-branch \slmT{degradation} encoder in DASAE.
    }
    \label{mutil_branch}
\end{figure}

\subsubsection{Multi-branch {degradation} encoder with perception gating}

Conventional single encoders often struggle to encompass diverse degradation patterns simultaneously. Conversely, employing multiple encoders to decouple degradation features in parallel offers inherent advantages in achieving task discrimination and feature adaptation within multi-degradation scenarios. Thus, we design a multi-branch encoder gating mechanism, where distinct branches are dedicated to extracting features specific to different degradation types. As shown in Fig.~\ref{mutil_branch}, the encoder comprises $k$ sub-encoders, each sub-encoder comprises three downsampling blocks, and each block consists of two convolutional layers and a downsampling layer. The DPG assigns weights to these sub-encoders based on the input features, thereby fusing the multi-path features. 

Given a degraded image $\mathbf{I} \in \mathbb{R}^{H \times W \times 3}$, each sub-encoder first extracts its corresponding degradation features. These features are then weighted by the perception confidence score $\mathbf{w} = [w_1, \cdots, w_k]$ generated by the gate. This process can be formulated as:
\begin{equation}
\begin{aligned}
&\{\mathbf{X}_i^1,\mathbf{X}_i^2, \mathbf{X}^3_i,\mathbf{X}^4_i\}= \operatorname{SubEncoder}_i(\mathbf{I}), \quad i=1,\cdots, k,  \\
&\mathbf{w} = [w_1, \cdots, w_k] = \operatorname{Softmax}(\mathcal{SE}(\mathbf{X}_1^4,\cdots,\mathbf{X}_k^4)), 
\label{eq:MOE-task}
\end{aligned}
\end{equation}
where $\operatorname{SubEncoder}_i(\cdot)$ represents the $i$-th sub-encoder and $\mathbf{w}$ is the perception confidence score. $\mathbf{X}_i^m (m \in \{1, 2, 3\})$ is the feature extracted by the convolutional layer of the $m\text{-th}$ downsampling block, and $\mathbf{X}_i^4$ is the feature extracted by the downsampling layer of the third downsampling block. 
$\mathcal{SE}(\cdot)$ is a Squeeze-and-Excitation block~\cite{hu2018squeeze}. 
The perception confidence score $\mathbf{w}$ generated by each sub-encoder will be integrated into the DPG-guided feature aggregation for feature fusion.

To recover spatial details and align features with the degradation pattern, we use a dynamic aggregation strategy guided by a degradation perception gate.
In contrast to static skip connections, our mechanism utilizes the perception confidence score $w_i$ from the DPG as a navigational signal to dynamically synthesize a unified latent representation from the $k$ \rev{sub-encoders}. Specifically, for both the shallow features at the $m$-th level and the final deep semantic features of {the} encoder, we perform a weighted projection:
\begin{equation}
\begin{aligned}
&\mathbf{X}_{skip}^m = \sum_{i=1}^k w_i \cdot \mathbf{X}^m_i, \quad m \in \{1,2,3\}, \\
&\mathbf{X}_{task} = \sum_{i=1}^k w_i \cdot \mathbf{X}_i^4,
\end{aligned}
\end{equation}
where $\mathbf{X}_{skip}^m$ denotes the aggregated adaptive skip features, which {help suppress} noise interference irrelevant to the current degradation. $\mathbf{X}_{task}$ represents the deep semantic features fused via the soft routing mechanism of the DPG.
 
Subsequently, $\mathbf{X}_{task}$ is fed into an Atrous Spatial Pyramid Pooling (ASPP) module~\cite{chen2017aspp} to generate $\mathbf{X}_{ASPP}$, further enriching the multi-scale contextual representations of the deep features. 

During the subsequent decoding phase, the fused shallow features $\mathbf{X}_{skip}^m$ are integrated into the structure decoder. The structure decoder comprises three upsampling blocks. Each block employs transposed convolution to upsample features from the preceding level, thereby restoring spatial resolution. These upsampled features are then concatenated with $\mathbf{X}_{skip}^m$ along the channel dimension and subsequently fed into a convolutional module to facilitate feature fusion:
\begin{equation}
\begin{aligned}
    &\mathbf{X}_{dec}^m = \mathcal{F}_{dec}^m \left( \left[\mathcal{U}(\mathbf{X}_{in}^m), \mathbf{X}_{skip}^m\right] \right), \quad m \in \{1, 2, 3\}, \\
    &\mathbf{X}_{in}^m = \begin{cases} 
        \mathbf{X}_{dec}^{m+1}, & m \in \{1, 2\}; \\ 
        \mathbf{X}_{ASPP}, & m=3,
    \end{cases} 
\end{aligned}
\end{equation}
where $[\cdot, \cdot]$ denotes the concatenation operation, $\mathcal{U}(\cdot)$ denotes the upsampling operation, and $\mathcal{F}_{dec}^m(\cdot)$ represents the feature fusion with two ConvBNReLU operations in the $m$-th upsampling block. Finally, $\mathbf{X}_{dec}^1$ is processed via a $1 \times 1$ convolution followed by a sigmoid activation function to generate the predicted structure map $\mathbf{I}_M$.

\subsubsection{Degradation-informed routing regularization}

To \rev{prevent the gating block from succumbing to mode collapse~\cite{chi2022representationMOE_ts} during training and encourage degradation-discriminative routing}, we introduce a degradation-informed routing regularization.

We formulate external degradation class labels as a distribution of prior knowledge $\mathbf{y} \in \{0,1\}^k$. Our objective is to minimize the Kullback-Leibler (KL) divergence between the generated perception confidence score distribution $\mathbf{w}$ and the target prior distribution $\mathbf{y}$. 
\slmT{In this way, degradation labels provide an auxiliary training signal for regularizing the latent routing distribution, encouraging the predicted routing scores to align with the degradation-label prior.}
Formally, this regularization term is defined as:
\begin{equation}
\begin{aligned}
\mathcal{L}_{reg} &= \mathcal{D}_{KL}(\mathbf{y}||\mathbf{w}) = \sum_{i=1}^{k} y_{i} \log \frac{y_{i}}{w_{i}}  \\
&= \underbrace{\sum_{i=1}^{k} y_{i} \log y_{i}}_{\text{Constant}} - \underbrace{\sum_{i=1}^{k} y_{i} \log w_{i}}_{\text{Cross-entropy}}.
\end{aligned}
\end{equation}
By imposing this explicit constraint, the DPG is \slmT{encouraged} to learn discriminative degradation features, thereby \slmT{improving} consistency between the routing logic and the input degradation patterns. Since the target distribution $\mathbf{y}$ is deterministic, its entropy $\sum_{i=1}^{k}y_{i}\log y_{i}$ is a constant. Consequently, minimizing the KL divergence is mathematically strictly equivalent to minimizing the cross-entropy loss.

\begin{figure}[t]
    \centering
    \includegraphics[width=1\columnwidth]{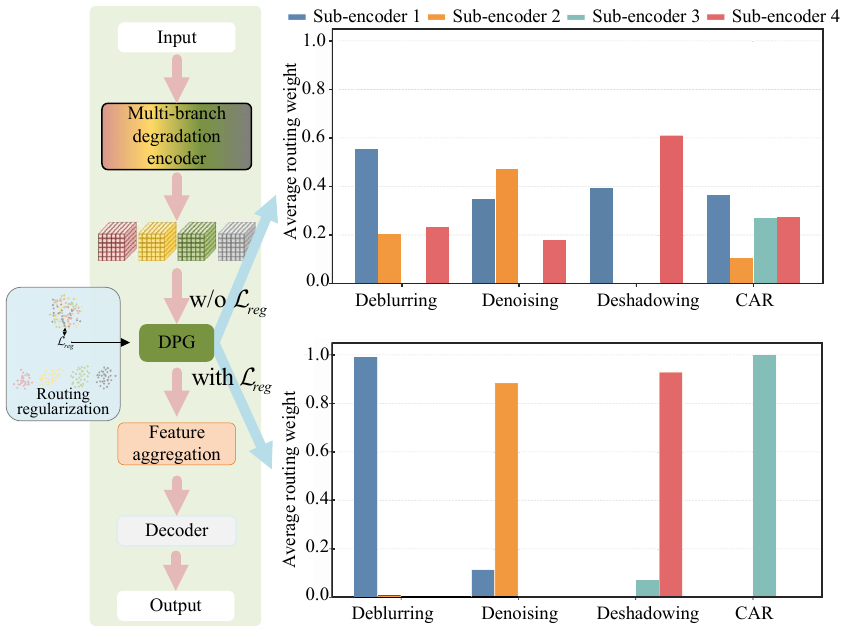}
    \caption{Visualization of degradation-informed routing regularization. We present the average routing weights across four semantic constraints: deblurring, denoising, CAR, and deshadowing. 
    The visualization demonstrates that our router learns a \slmT{more sparse and class-selective routing} policy, \slmT{effectively separating distinct degradation patterns}. Each sub-encoder is {primarily} focused on a specific task
    (\eg sub-encoder 1 for deblurring, sub-encoder 4 for CAR), verifying the effectiveness of the proposed routing mechanism in distinguishing task semantics.}
    \label{fig:routing_weight}
    \end{figure}

We visualize the average routing weight distribution of the DPG within the DASAE module across four tasks in Fig.~\ref{fig:routing_weight}. Without the routing regularization term $\mathcal{L}_{reg}$, the gating block exhibits 
\slmT{less separable routing behavior.}
For specific degradation types, the weight distribution appears relatively dispersed, with multiple encoders being activated simultaneously and lacking significant weight distinctiveness. This failure to separate degradation patterns causes the encoders to learn similar features, which undermines the specialization of individual encoders. With the introduction of $\mathcal{L}_{reg}$, the routing weights exhibit high sparsity and orthogonality. Each degradation task almost exclusively activates a specific dominant encoder. This one-to-one mapping relationship indicates that $\mathcal{L}_{reg}$ successfully compels the DPG to learn discriminative degradation features for different degradation patterns.

\subsubsection{Loss function of DASAE}
To achieve robust performance in terms of both boundary details and overall structure, we formulate a composite loss function. We employ binary cross-entropy loss $\mathcal{L}_{bce}$ to enhance pixel-level accuracy. To address the inherent class imbalance between the sparse text foreground and the dominant background, we incorporate the Dice loss function ($\mathcal{L}_{dice}$)~\cite{sudre2017loss_dice}. As a region-based metric, it has been proven effective in handling segmentation tasks characterized by class imbalance, achieving this by normalizing according to the size of the target region. However, while $\mathcal{L}_{dice}$ ensures region overlap, preserving the fine structural integrity of character strokes necessitates precise boundary localization. To this end, we introduce the boundary loss ($\mathcal{L}_{bd}$), which minimizes the distance between predicted contours and GT contours~\cite{kervadec2019loss_bd}. By explicitly penalizing boundary deviations, this loss complements the region-based metrics, thereby encouraging the network to generate sharp text structures. The total loss is formulated as follows:
\begin{equation}
\begin{aligned}
\mathcal{L} =&
\lambda_{bce} \mathcal{L}_{bce} +
\lambda_{dice} \mathcal{L}_{dice} +
\lambda_{bd} \mathcal{L}_{bd}+\mathcal{L}_{reg}\\
  =
  & \underbrace{ - \frac{\lambda_{bce}}{N} \sum_{j=1}^{N} [g_j \log(p_j) + (1 - g_j) \log(1 - p_j)] }_{\text{BCE loss}} \\
 & + \underbrace{ \lambda_{dice} \left( 1 - \frac{2 \sum_{j=1}^{N} p_j g_j }{\sum_{j=1}^{N} p_j + \sum_{j=1}^{N} g_j } \right) }_{\text{Dice loss}}  \\
& + \underbrace{ \frac{\lambda_{bd}}{N} \sum_{j=1}^{N} \phi_{GT(j)} p_j }_{\text{Boundary loss}} + \underbrace{ \mathcal{L}_{reg}}_{\text{Routing regularization}}, 
\end{aligned}
\end{equation}
where $N$ denotes the total number of pixels in the output image, $\lambda_{bce}$, $\lambda_{dice}$, and $\lambda_{bd}$ are weights to balance the loss functions. Let $g_j \in \{0, 1\}$ and $p_j \in [0, 1]$ denote the GT label and the predicted probability for the $j$-th pixel, respectively. $\phi_{GT(j)}$ denotes the level set function derived from the GT, which corresponds to the signed distance map, \ie the signed distance from pixel $j$ to the nearest boundary contour.

\subsection{Structure-guided wavelet interaction module (SWIM)}

Degradation in document images typically involves loss of high-frequency information and structural discontinuities, imposing stringent demands on the restoration of structures and strokes. Local spatial features alone fail to capture frequency attenuation and global structure, limiting the reconstruction of details. While high-frequency components inherently carry structural information, directly utilizing them from degraded images renders the process susceptible to interference from noise and pseudo-textures. The direct application of predicted structure maps within the spatial domain introduces substantial irrelevant binary redundancy, which can disrupt feature extraction.

As illustrated in Fig.~\ref{model1}~(c), {to effectively integrate structural priors with frequency components, we propose a structure-guided wavelet interaction mechanism. This mechanism is implemented as the structure-guided wavelet interaction module (SWIM).} Instead of simple feature concatenation, SWIM is conceptually formulated to bridge the semantic gap between spatial layouts and spectral cues. By establishing a collaborative interaction in the wavelet domain, it enables the network to explicitly perceive degradation patterns from spectral signatures while robustly guiding the recovery of fine-grained structural details.
Specifically, we employ the Haar wavelet basis~\cite{mallat1989wavelet} to perform wavelet decomposition on the input image $\mathbf{I}$, obtaining a low-frequency component (LL) and three high-frequency components (HL, LH, HH). The low-frequency component is mapped to features $\mathbf{F}_L \in \mathbb{R}^{H \times W \times C}$, while the three high-frequency components are mapped to $\mathbf{F}_H^{rgb} \in \mathbb{R}^{H \times W \times C}$. Meanwhile, the high-frequency components obtained from the wavelet decomposition of the predicted structure map $\mathbf{I}_M$ are mapped to $\mathbf{F}_H^{st} \in \mathbb{R}^{H \times W \times C}$. The process is expressed as:
\begin{equation}
\begin{aligned}
    &(\mathbf{I}_{LL}, \mathbf{I}_{HL}, \mathbf{I}_{LH}, \mathbf{I}_{HH}) = \mathcal{W}_{Haar}(\mathbf{I}), \\
    &\mathbf{F}_L = \operatorname{Conv}_{3\times3}(\mathbf{I}_{LL}),  \\
    &\mathbf{F}_H^{rgb} = \operatorname{Conv}_{3\times3}([\mathbf{I}_{HL}, \mathbf{I}_{LH}, \mathbf{I}_{HH}]),  \\
    &(\mathbf{M}_{LL}, \mathbf{M}_{HL}, \mathbf{M}_{LH}, \mathbf{M}_{HH}) = \mathcal{W}_{Haar}(\mathbf{I}_M), \\
    &\mathbf{F}_H^{st} = \operatorname{Conv}_{3\times3}([\mathbf{M}_{HL}, \mathbf{M}_{LH}, \mathbf{M}_{HH}]),
\end{aligned}
\end{equation}
where $[\cdot,\cdot,\cdot]$ denotes the concatenation operation, $\operatorname{Conv}_{3\times3}(\cdot)$ denotes a $3\times3$ convolutional layer, and $\mathcal{W}_{Haar}(\cdot)$ signifies the wavelet transform utilizing the Haar wavelet basis for frequency-domain feature extraction.

To inject frequency priors into the semantic space, we employ cross-attention~\cite{vaswani2017attention,cui2024adair}. By utilizing the spatial distribution of frequency features as indices, we retrieve and aggregate relevant semantic contexts from $\mathbf{Y}$, which represents the spatial features extracted by the Transformer block, thereby generating frequency-aware features $\mathbf{F}_L', \mathbf{F}_H'^{rgb}$, and $\mathbf{F}_H'^{st}$. This operation enables the model to extract multi-scale semantic information aligned with the attentional focus of distinct frequency components.

To balance the influence of image textures and predicted structures, we design a spatial gating module $\operatorname{G}(\cdot)$. Taking the raw frequency features $\mathbf{F}_H^{rgb}$ and $\mathbf{F}_H^{st}$ as input, $\operatorname{G}(\cdot)$ leverages the physical structural information from both low and high frequencies to learn pixel-wise weight maps $\mathbf{W}_1, \mathbf{W}_2 \in \mathbb{R}^{H \times W \times 1}$. These weight maps are subsequently employed to modulate the frequency-aware features $\mathbf{F}_H'^{rgb}$ and $\mathbf{F}_H'^{st}$ derived in the preceding step:
\begin{equation}
\begin{aligned}
    & \mathbf{F}'_L = \operatorname{CA}(\mathbf{F}_L, \mathbf{Y}, \mathbf{Y}), \\
    & \mathbf{F}'^{rgb}_H = \operatorname{CA}(\mathbf{F}_H^{rgb}, \mathbf{Y}, \mathbf{Y}), \\
    & \mathbf{F}'^{st}_H = \operatorname{CA}(\mathbf{F}_H^{st}, \mathbf{Y}, \mathbf{Y}), \\
    & \mathbf{W}_1, \mathbf{W}_2 = \operatorname{G}([\mathbf{F}_H^{rgb}, \mathbf{F}_H^{st}]), \\
    & \mathbf{F}'_H = \operatorname{Conv}_{1\times1}(\mathbf{W}_1 \odot \mathbf{F}'^{rgb}_H + \mathbf{W}_2 \odot \mathbf{F}'^{st}_H),
\end{aligned}
\end{equation}
where $\operatorname{CA}(\mathbf{Q},\mathbf{K},\mathbf{V})$ denotes the cross-attention module. $\mathbf{Y}$ serves as the source for both keys ($\mathbf{K}$) and values ($\mathbf{V}$), while the frequency branch features $\mathbf{F}_L$, $\mathbf{F}_H^{rgb}$, and $\mathbf{F}_H^{st}$ independently act as the queries ($\mathbf{Q}$). This design is aimed at preserving the inherent spatial resolution and fine-grained structures of the frequency features. By explicitly injecting the global semantic context from $\mathbf{Y}$ into each frequency component via the attention mechanism\cite{vaswani2017attention}, it compensates for the semantic deficiency inherent in low-level visual features. The module $\operatorname{G}(\cdot)$ comprises two convolutional layers followed by ReLU activation, culminating in a Softmax function to generate complementary spatial attention maps. Leveraging the raw frequency structure as a prior, this gating mechanism adaptively adjusts the weights assigned to the high-frequency perceptual components derived from the predicted structure map.

We feed the fused high-frequency perceptual features $\mathbf{F}_H'$ into the CFAM module, where they are adaptively modulated by the low-frequency perceptual features $\mathbf{F}_L'$, producing the enhanced features $\mathbf{F}_{out} \in \mathbb{R}^{H \times W \times C}$:
\begin{equation}
    \mathbf{F}_{out} = \operatorname{CFAM}(\mathbf{F}_H', \mathbf{F}_L').
\end{equation}
We then employ the cross-attention module once again to re-integrate the enhanced features $\mathbf{F}_{out}$ into the backbone semantic features $\mathbf{Y}$, thereby updating the feature representation:
\begin{equation}
    \mathbf{Y}' = \operatorname{CA}(\mathbf{Y}, \mathbf{F}_{out}, \mathbf{F}_{out}),
\end{equation}
where we utilize $\mathbf{Y}$ as the query ($\mathbf{Q}$), while $\mathbf{F}_{out}$, which encapsulates rich high-frequency information, serves as both the key ($\mathbf{K}$) and value ($\mathbf{V}$). This design is intended to retrieve and aggregate relevant spatial texture details from the refined frequency-domain features, guided by the backbone semantic features. By propagating the enhanced high-frequency information back into the semantic space, it enriches the detailed representation of features while preserving semantic consistency.

\textbf{Cross-frequency adaptive modulation (CFAM).}
Embedded within SWIM, CFAM is designed to further refine feature representations. 
It is designed to calibrate high-frequency details leveraging the global structural consistency of low-frequency priors. 

We propose a low-frequency guided dual modulation mechanism.
The high-frequency feature $\mathbf{F}_H'$ undergoes a local enhancement via a $1\times1$ convolution and two depthwise separable convolutions to obtain $\mathbf{F}_H''$. The low-frequency branch $\mathbf{F}_L'$ branches into two parallel paths to generate modulation parameters. Spatially, it encodes long-range structural contexts to produce an attention map $\mathbf{G}_{map}$, which effectively suppresses noise in high-frequency regions. Meanwhile, in the channel dimension, global statistics of $\mathbf{F}_L'$ are aggregated to generate affine transformation factors ($\boldsymbol{\gamma}, \boldsymbol{\beta}$) for feature recalibration.
Finally, $\mathbf{F}_H''$ is spatially gated and channel-wise modulated to achieve restoration:
\begin{equation}
\tilde{\mathbf{F}}_H = \boldsymbol{\gamma} \odot (\mathbf{F}_H'' \odot \mathbf{G}_{map}) + \boldsymbol{\beta},
\end{equation}
where $\odot$ denotes element-wise multiplication. The calibrated $\tilde{\mathbf{F}}_H$ is then fused with the  $\mathbf{F}_L'$ to produce the output.

\subsection{Loss function of restoration network}
Following the previous research~\cite{cui2024adair,zamir2022restormer},
for the overall restoration network, we adopt the $\ell_1$ norm as the training loss function $\mathcal{L}_{\text{rec}}$ to quantify the pixel-wise discrepancy between the restored image $\hat{\mathbf{I}}$ and the GT image $\tilde{\mathbf{I}}$:
\begin{equation}\mathcal{L}_{\text{rec}} = \|\tilde{\mathbf{I}} - \hat{\mathbf{I}}\|_1.\end{equation}
This loss function is simple yet effective, encouraging the model to generate restoration results that are closer to the target image within the pixel space.

\section{Experiments}

\subsection{Datasets} 

The datasets used in this paper are summarized in Table~\ref{tab:datasets_summary}.

\begin{table}[t]
\centering
\slmT{
\caption{Summary of datasets used in this paper.}
\label{tab:datasets_summary}
\footnotesize 
\setlength{\tabcolsep}{2pt} 
\resizebox{\columnwidth}{!}{
\begin{tabular}{@{}lr|lr|c@{}}
\toprule
Source dataset & \#Images &  Derived dataset & \#Images & Usage \\
\midrule

\multicolumn{5}{@{}l}{{Deblurring task}} \\
TDD training set~\cite{hradivs2015TDD} & 66,000 & TDD training set & 40,000 & Training \\
TDD test set~\cite{hradivs2015TDD} & 1,600 & TDD test set & 1,600 & Test \\
FUNSD~\cite{jaume2019funsd} & 199 & FUNSD$_{blur}$ & 199 & Test \\
BCSD~\cite{khan2021BCSD} & 158 & BCSD$_{blur}$ & 158 & Test \\
\midrule

\multicolumn{5}{@{}l}{{Denoising task}} \\
TDD training set~\cite{hradivs2015TDD} & 66,000 & TDD$_{noise}$ training set & 40,000 & Training \\
TDD test set~\cite{hradivs2015TDD} & 1,600 & TDD$_{noise}$ test set & 12,800 & Test \\
FUNSD~\cite{jaume2019funsd} & 199 & FUNSD$_{noise}$ & 1,592 & Test \\
BCSD~\cite{khan2021BCSD} & 158 & BCSD$_{noise}$ & 1,264 & Test \\
\midrule

\multicolumn{5}{@{}l}{{Compression artifacts reduction task}} \\
TDD training set~\cite{hradivs2015TDD} & 66,000 & TDD$_{CAR}$ training set & 40,000 & Training \\
TDD test set~\cite{hradivs2015TDD} & 1,600 & $\text{TDD}_{CAR}$ test set & 17,600 & Test \\
FUNSD~\cite{jaume2019funsd} & 199 & $\text{FUNSD}_{CAR}$ & 796 & Test \\
BCSD~\cite{khan2021BCSD} & 158 & $\text{BCSD}_{CAR}$ & 632 & Test \\
\midrule

\multicolumn{5}{@{}l}{{Deshadowing task}} \\
FSDSRD~\cite{matsuo2022FSD} & 14,200 & FSDSRD & 14,200 & Training \\
RDD training set~\cite{zhang2023RDD} & 4,371 & RDD training data & 4,371 & Training \\
RDD test set~\cite{zhang2023RDD} & 545 & RDD test set & 545 & Test \\
OSR~\cite{wang2020OSR} & 237 & OSR & 237 & Test \\
Jung et al.~\cite{jung2018jun} & 87 & Jung et al. & 87 & Test \\
\bottomrule
\end{tabular}
}}
\end{table}

\textbf{Deblurring.} For the text deblurring task, we utilized the text deblurring dataset (TDD)~\cite{hradivs2015TDD} as our benchmark. This dataset comprises 66,000 blurry-clean pairs for training and 1,600 samples for testing. Following the convention of recent literature~\cite{zhang2024docres}, we randomly sampled 40,000 pairs from the TDD training set to train our model and employed its full testing set for evaluation. To assess the generalization capability of the model on unseen document domains, we further tested on two additional datasets with distinct styles: FUNSD~\cite{jaume2019funsd} and BCSD~\cite{khan2021BCSD}, which contain 199 and 158 GT images, respectively. We synthesized blurry versions using the blur kernels provided by TDD, thereby constructing two extra test sets denoted as FUNSD$_{blur}$ and BCSD$_{blur}$.

\textbf{Denoising.} For the text denoising task, we constructed a noisy document dataset TDD$_{noise}$ based on TDD. For training, we randomly selected 40,000 GT images from the TDD training set and corrupted each image with additive zero-mean Gaussian noise of varying intensities. The standard deviation $\sigma \in \{0, 1, 2, \ldots, 35\}$ was uniformly sampled to cover a wide range of scenarios, from minor interference to severe contamination. The noise was added independently to each RGB channel. For testing, to systematically measure the denoising performance under different noise levels, we constructed hierarchical noise test sets based on the TDD test dataset, FUNSD, and BCSD, denoted as TDD$_{noise}$, FUNSD$_{noise}$, and BCSD$_{noise}$, respectively. Specifically, we added independent and identically distributed (i.i.d.) zero-mean Gaussian noise with $\sigma \in \{0, 5, 10, \ldots, 35\}$ to each GT image.

\textbf{Compression artifacts reduction (CAR).} For the compression artifacts reduction task, we selected 40,000 high-quality GT samples from the TDD training set. By applying JPEG encoding to each image with a randomly selected quality factor $q \in [2, 12]$, we simulated compression distortions ranging from mild to severe. This process generated a rich set of degradation-clean pairs to train the model for robust recovery across varying compression intensities. 
For testing, we constructed the $\text{TDD}_{CAR}$ set with $q \in [2, 12]$. Additionally, to evaluate robustness under {severe compression scenarios}, we generated $\text{BCSD}_{CAR}$ and $\text{FUNSD}_{CAR}$ using a restricted range of $q \in [2, 5]$, as compression artifacts become less challenging at higher quality factors.

\textbf{Deshadowing.} The FSDSRD dataset~\cite{matsuo2022FSD} and the RDD training dataset~\cite{zhang2023RDD} were combined as the training set for our deshadowing task. FSDSRD consists of 14,200 synthetic images, while the RDD training dataset consists of 4,371 real-world images. The RDD test dataset consisting of 545 {real-world} images was employed directly as our test dataset. To evaluate the generalization performance on cross-domain data, {we further tested on OSR~\cite{wang2020OSR} and Jung et al.'s dataset~\cite{jung2018jun}, which contains 237 {synthetic} images and 87 {real-world} images, respectively.}

\textbf{GT structure maps.} For the FSDSRD dataset, the provided binary structure annotations were employed as GT  structure maps. For other datasets, GT structure maps were generated by applying Sauvola's adaptive thresholding binarization~\cite{sauvola2000adaptive} to GT images.

\begin{table*}[t]
\caption{Quantitative comparison with SOTA document restoration methods on various degradation datasets. Our DocPure achieves {the best overall} performance. \textbf{Bold} indicates the best and \underline{underlined} the second best. } 
\centering
\label{tab:1}
\setlength{\tabcolsep}{2pt}
\resizebox{\textwidth}{!}{%
\begin{tabular}{c|cc|ccccc|c}
\hline
Task &
  Dataset &
  \multicolumn{1}{c|}{Metric} &
  \begin{tabular}[c]{@{}c@{}}DE-GAN~\cite{souibgui2020gan}\\ TPAMI'20\end{tabular} &
  \begin{tabular}[c]{@{}c@{}}DocDiff~\cite{yang2023docdiff}\\ ACM MM'23\end{tabular} &
  \begin{tabular}[c]{@{}c@{}}DocStain~\cite{li2025high}\\ WACV'25\end{tabular} &
  \begin{tabular}[c]{@{}c@{}}DocNLC~\cite{wang2024docnlc}\\ AAAI'24\end{tabular} &
  \begin{tabular}[c]{@{}c@{}}DocRes~\cite{zhang2024docres}\\ CVPR'24\end{tabular} &
  \begin{tabular}[c]{@{}c@{}}DocPure\\Ours\end{tabular} \\ \hline
\multicolumn{1}{c|}{Deblurring} &
  TDD &
  \multicolumn{1}{c|}{\begin{tabular}[c]{@{}c@{}}PSNR / SSIM $\uparrow$\\ Char. Acc. / \slmT{Word Acc.} (Paddle) (\%) $\uparrow$\\ \slmT{Char. Acc. / Word Acc. (Nougat) (\%) $\uparrow$}\end{tabular}} &
  \begin{tabular}[c]{@{}c@{}}22.24 / 0.9226\\ 61.59 / \slmT{29.06}\\ \slmT{32.69 / 28.56}\end{tabular} &
  \begin{tabular}[c]{@{}c@{}}24.00 / 0.9559\\ 77.89 / \slmT{44.93}\\ \slmT{37.70 / 29.38}\end{tabular} &
  \begin{tabular}[c]{@{}c@{}}\underline{30.05} / \underline{0.9861}\\ \underline{85.94} / \slmT{\underline{60.34}}\\ \slmT{\underline{50.45} / \underline{43.05}}\end{tabular} &
  \begin{tabular}[c]{@{}c@{}}16.56 / 0.8061\\ 34.21 / \slmT{9.74}\\ \slmT{7.06 / 10.15}\end{tabular} &
  \begin{tabular}[c]{@{}c@{}}28.34 / 0.9815\\ 83.22 / \slmT{55.66}\\ \slmT{47.30 / 37.74}\end{tabular} &
  \textbf{\begin{tabular}[c]{@{}c@{}}32.55 / 0.9912\\ 87.25 / \slmT{\textbf{61.27}}\\ \slmT{56.31 / \textbf{51.41}}\end{tabular}} \\ \hline
\multicolumn{1}{c|}{Denoising} &
  TDD$_{noise}$ &
  \multicolumn{1}{c|}{\begin{tabular}[c]{@{}c@{}}PSNR / SSIM $\uparrow$\\ Char. Acc. / \slmT{Word Acc.} (Paddle) (\%) $\uparrow$\\ \slmT{Char. Acc. / Word Acc. (Nougat) (\%) $\uparrow$}\end{tabular}} &
  \begin{tabular}[c]{@{}c@{}}38.32 / 0.9947\\ 87.62 / \slmT{64.85}\\ \slmT{68.16 / 64.93}\end{tabular} &
  \begin{tabular}[c]{@{}c@{}}37.03 / 0.9912\\ 86.62 / \slmT{62.53}\\ \slmT{66.80 / 63.46}\end{tabular} &
  \textbf{\begin{tabular}[c]{@{}c@{}}43.07 / 0.9987\\ 92.41 / \slmT{\textbf{74.15}}\\ \slmT{74.95 / \textbf{72.39}}\end{tabular}} &
  \begin{tabular}[c]{@{}c@{}}19.98 / 0.9047\\ 63.66 / \slmT{19.81}\\ \slmT{4.67 / 12.52}\end{tabular} &
  \begin{tabular}[c]{@{}c@{}}37.91 / 0.9947\\ 83.89 / \slmT{69.13}\\ \slmT{64.84 / 61.37}\end{tabular} &
  \begin{tabular}[c]{@{}c@{}}\underline{41.26} / \underline{0.9973}\\ \underline{91.29} / \slmT{\underline{71.73}}\\ \slmT{\underline{72.30} / \underline{69.66}}\end{tabular} \\ \hline
\multicolumn{1}{c|}{CAR} &
  TDD$_{CAR}$ &
  \multicolumn{1}{c|}{\begin{tabular}[c]{@{}c@{}}PSNR / SSIM $\uparrow$\\ Char. Acc. / \slmT{Word Acc.} (Paddle) (\%) $\uparrow$\\ \slmT{Char. Acc. / Word Acc. (Nougat) (\%) $\uparrow$}\end{tabular}} &
  \begin{tabular}[c]{@{}c@{}}18.71 / 0.8871\\ 70.32 / \slmT{28.88}\\ \slmT{22.89 / 16.20}\end{tabular} &
  \begin{tabular}[c]{@{}c@{}}27.41 / 0.9778\\ 83.89 / \slmT{\underline{54.56}}\\ \slmT{43.11 / 38.00}\end{tabular} &
  \begin{tabular}[c]{@{}c@{}}\underline{29.37} / 0.9710\\ \underline{85.69} / \slmT{53.73}\\ \slmT{\underline{45.53} / \underline{46.94}}\end{tabular} &
  \begin{tabular}[c]{@{}c@{}}18.82 / 0.8837\\ 55.34 / \slmT{16.53}\\ \slmT{4.55 / 11.20}\end{tabular} &
  \begin{tabular}[c]{@{}c@{}}28.69 / \underline{0.9804}\\ 84.19 / \slmT{50.90}\\ \slmT{42.34 / 37.15}\end{tabular} &
  \textbf{\begin{tabular}[c]{@{}c@{}}31.34 / 0.9893\\ 86.43 / \slmT{\textbf{59.86}}\\ \slmT{53.40 / \textbf{48.96}}\end{tabular}} \\ \hline
\multicolumn{1}{c|}{Deshadowing} &
  RDD &
  \multicolumn{1}{c|}{\begin{tabular}[c]{@{}c@{}}PSNR / SSIM $\uparrow$\\ Char. Acc. / \slmT{Word Acc.} (Paddle) (\%) $\uparrow$\\ \slmT{Char. Acc. / Word Acc. (Nougat) (\%) $\uparrow$}\end{tabular}} &
  \begin{tabular}[c]{@{}c@{}}32.02 / 0.9223\\ 53.65 / \slmT{44.31}\\ \slmT{83.59 / 84.13}\end{tabular} &
  \begin{tabular}[c]{@{}c@{}}27.88 / 0.7933\\ 47.85 / \slmT{40.48}\\ \slmT{74.67 / 76.06}\end{tabular} &
  \begin{tabular}[c]{@{}c@{}}32.82 / 0.9281\\ 57.32 / \slmT{47.41}\\ \slmT{83.61 / 83.88}\end{tabular} &
  \begin{tabular}[c]{@{}c@{}}- / -\\ - / \slmT{-}\\ \slmT{- / -}\end{tabular} &
  \begin{tabular}[c]{@{}c@{}}\textbf{34.35} / \underline{0.9528}\\ \underline{64.42} / \slmT{\underline{53.79}}\\ \slmT{\underline{85.54} / \textbf{90.21}}\end{tabular} &
  \begin{tabular}[c]{@{}c@{}}\underline{33.31} / \textbf{0.9532}\\ \textbf{65.69} / \slmT{\textbf{56.23}}\\ \slmT{\textbf{87.85} / \underline{89.35}}\end{tabular} \\ \hline
  &&\slmT{Average rank $\downarrow$} &
  \slmT{4.31} &
  \slmT{4.17} &
  \slmT{\underline{2.17}} &
  \slmT{5.96} &
  \slmT{3.06} &
  \slmT{\textbf{1.33}} \\ \hline
\end{tabular}
}
\end{table*}

\begin{figure*}[t]
    \centering
    \includegraphics[width=\textwidth]{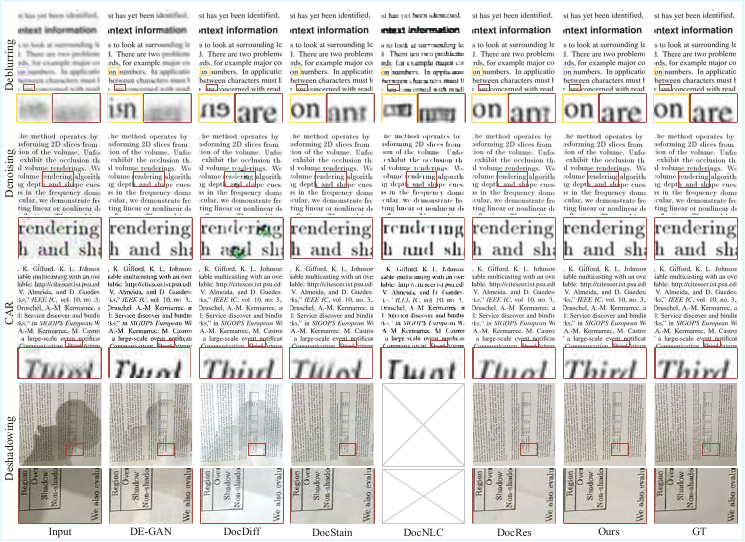}
    \caption{Qualitative comparison across four restoration tasks: deblurring, denoising, CAR, and deshadowing, evaluated on the TDD, TDD$_{CAR}$, TDD$_{noise}$, and RDD datasets, respectively. As evidenced by the results on TDD and TDD$_{CAR}$, our method recovers {clearer} text boundaries and rectifies distorted strokes. This {enhanced} structural fidelity is {consistent with the role of} the DASAE, which predicts structure maps and injects high-frequency priors to guide the reconstruction. For denoising on TDD$_{noise}$ and deshadowing on RDD, DocPure achieves clean background separation while preserving intricate details.}
    \label{compare1}
\end{figure*}

\subsection{Implementation details}

To comprehensively evaluate the performance of DocPure, we conducted extensive benchmarking against a suite of representative SOTA methods, including general-task restoration models (DE-GAN~\cite{souibgui2020gan}, DocDiff~\cite{yang2023docdiff}, and DocStain~\cite{li2025high}), and all-in-one restoration models (DocNLC~\cite{wang2024docnlc} and DocRes~\cite{zhang2024docres}). To ensure a fair comparison, all compared models were retrained using their official codes and default configurations.
General-task models were trained on individual task-specific training datasets according to restoration tasks: TDD (deblurring), TDD$_{noise}$ (denoising), TDD$_{CAR}$ (CAR), FSDSRD and RDD (deshadowing). All-in-one models (including our DocPure) were trained across all training datasets. Consequently, multiple models were trained for each general-task method, while only one model \slmT{was} trained for each all-in-one method. All compared models were trained on NVIDIA GeForce {RTX} 4090 GPUs.

The entire training process of the proposed DocPure was decoupled into two stages: the DASAE network was trained first, followed by the restoration network.

\textbf{Training of our DASAE network.} We employed the AdamW optimizer~\cite{loshchilov2017adamw} for parameter optimization, with hyperparameters set to $\beta_1 = 0.9$ and $\beta_2 = 0.999$. The learning rate was maintained constant at $1 \times 10^{-4}$ throughout the training, with a batch size of 32. For the loss function, we set the weights to $\lambda_{bce} = 0.3$, $\lambda_{dice} = 0.5$, and $\lambda_{bd} = 0.35$. 
In this paper, we have implemented our method with four tasks, \ie deblurring, denoising, CAR, and deshadowing. Since each sub-encoder is designed to correspond to a restoration task, we set the number of sub-encoders to $k=4$. In order to improve the training efficiency, we employed a progressive training strategy. First, we conducted pre-training on the deblurring task for 20 epochs by replacing the multi-branch \slmT{degradation} encoder with a single encoder. Then, we froze the structure decoder to 
serve as a shared 
\slmT{structural prediction}
constraint and proceeded to pre-train new encoders on the denoising, CAR, and deshadowing tasks for 20 epochs, respectively. Finally, we injected the above four pre-trained encoders into the sub-encoders of our multi-branch \slmT{degradation} encoder, unfroze the structure decoder, and trained across all training datasets for 50 epochs.
During this training phase, input images for all tasks were randomly cropped to a size of $300 \times 300$. {The training process took about 60 hours.}

\textbf{Training of our restoration network.} {The architecture of the restoration network employs a 4-level encoder-decoder structure, with a varying number of Transformer blocks at each level, specifically [4, 6, 6, 8] from level 1 to level 4.} We employed a batch size of 16 and trained for a total of 80 epochs. We utilized the AdamW optimizer, with hyperparameters set to $\beta_1 = 0.9$ and $\beta_2 = 0.999$. The learning rate was initialized at $2 \times 10^{-4}$ and gradually decayed to $1 \times 10^{-6}$ via the cosine annealing strategy~\cite{loshchilov2016cos}. The learning rate was adjusted at each iteration step to ensure a smoother optimization trajectory. During this phase, images from all tasks were randomly cropped into patches of size $128 \times 128$. {The training process took about 100 hours.}

\subsection{Metrics}

We employed the peak signal-to-noise ratio (PSNR), the structural similarity index (SSIM), and character accuracy as primary quantitative metrics to evaluate the model's performance in document image restoration tasks. PSNR measures the pixel-level reconstruction accuracy between the restored and GT images. SSIM comprehensively assesses the {structural} similarity between the restored and GT images. 
\slmT{Let $s_{res}$ and $s_{gt}$ be the text sequences extracted from the restored and GT images, the OCR character accuracy is defined as $1 - d_c(s_{res}, s_{gt}) /\max(|s_{res}|, |s_{gt}|)$, where $|\cdot|$ denotes the sequence length, {and $d_c(s_{res}, s_{gt})$ is the character-level edit distance measuring the difference between $s_{res}$ and $s_{gt}$. Similarly, by tokenizing the texts into word sequences $w_{res}$ and $w_{gt}$, the OCR word accuracy is computed as $1 - d_w(w_{res}, w_{gt}) /\max(|w_{res}|, |w_{gt}|)$, where $d_w(w_{res}, w_{gt})$ is the word-level edit distance between $w_{res}$ and $w_{gt}$.} Two OCR models (Paddle~\cite{cui2025paddleocrvlboostingmultilingualdocument} and Nougat~\cite{blecher2023nougat}) were tested to extract text sequences in our experiments.}

\begin{table*}[t]
\caption{Quantitative generalization comparison with SOTA document restoration methods on various degradation datasets. Generalization comparison was evaluated on test datasets with styles distinct from the training datasets. Our DocPure achieves {the best overall} performance. \textbf{Bold} indicates the best and \underline{underlined} the second best.}
\centering
\label{tab:compare_add}
\setlength{\tabcolsep}{2pt}
\resizebox{\textwidth}{!}{
\begin{tabular}{c|cc|ccccc|c}
\hline
Task &
  Dataset &
  Metric &
  \begin{tabular}[c]{@{}c@{}}DE-GAN~\cite{souibgui2020gan}\\ TPAMI'20\end{tabular} &
  \begin{tabular}[c]{@{}c@{}}DocDiff~\cite{yang2023docdiff}\\ ACM MM'23\end{tabular} &
  \begin{tabular}[c]{@{}c@{}}DocStain~\cite{li2025high}\\ WACV'25\end{tabular} &
  \begin{tabular}[c]{@{}c@{}}DocNLC~\cite{wang2024docnlc}\\ AAAI'24\end{tabular} &
  \begin{tabular}[c]{@{}c@{}}DocRes~\cite{zhang2024docres}\\ CVPR'24\end{tabular} &
  \begin{tabular}[c]{@{}c@{}}DocPure\\Ours\end{tabular} \\ \hline
\multirow{4}{*}{Deblurring} &
  FUNSD$_{blur}$ &
  \multicolumn{1}{c|}{\begin{tabular}[c]{@{}c@{}}PSNR / SSIM $\uparrow$\\ Char. Acc. / \slmT{Word Acc.} (Paddle) (\%) $\uparrow$\\ \slmT{Char. Acc. / Word Acc. (Nougat) (\%) $\uparrow$}\end{tabular}} &
  \begin{tabular}[c]{@{}c@{}}25.74 / 0.9666\\ 73.24 / \slmT{39.66}\\ \slmT{51.68 / 49.58}\end{tabular} &
  \begin{tabular}[c]{@{}c@{}}19.96 / 0.9083\\ 30.66 / \slmT{13.50}\\ \slmT{17.65 / 14.69}\end{tabular} &
  \begin{tabular}[c]{@{}c@{}}\textbf{34.11} / \textbf{0.9941}\\ \textbf{91.93} / \slmT{\textbf{67.46}}\\ \slmT{\textbf{76.62} / \textbf{78.78}}\end{tabular} &
  \begin{tabular}[c]{@{}c@{}}18.28 / 0.8683\\ 15.95 / \slmT{2.03}\\ \slmT{15.29 / 4.57}\end{tabular} &
  \begin{tabular}[c]{@{}c@{}}29.22 / 0.9852\\ 81.65 / \slmT{59.23}\\ \slmT{66.92 / 67.94}\end{tabular} &
  \begin{tabular}[c]{@{}c@{}}\underline{31.21} / \underline{0.9899}\\ \underline{89.79} / \slmT{\underline{65.45}}\\ \slmT{\underline{72.90} / \underline{75.03}}\end{tabular} \\ \cline{2-9}
\multicolumn{1}{c|}{} &
  BCSD$_{blur}$ &
  \multicolumn{1}{c|}{\begin{tabular}[c]{@{}c@{}}PSNR / SSIM $\uparrow$\\ Char. Acc. / \slmT{Word Acc.} (Paddle) (\%) $\uparrow$\\ \slmT{Char. Acc. / Word Acc. (Nougat) (\%) $\uparrow$}\end{tabular}} &
  \begin{tabular}[c]{@{}c@{}}23.65 / 0.8293\\ 78.61 / \slmT{47.70}\\ \slmT{82.26 / 87.19}\end{tabular} &
  \begin{tabular}[c]{@{}c@{}}19.78 / 0.6586\\ 67.25 / \slmT{24.11}\\ \slmT{81.59 / 81.80}\end{tabular} &
  \begin{tabular}[c]{@{}c@{}}28.11 / 0.8637\\ 85.14 / \slmT{\underline{52.40}}\\ \slmT{90.38 / \underline{90.78}}\end{tabular} &
  \begin{tabular}[c]{@{}c@{}}16.43 / 0.7186\\ 57.11 / \slmT{19.11}\\ \slmT{64.64 / 80.72}\end{tabular} &
  \begin{tabular}[c]{@{}c@{}}\underline{29.57} / \underline{0.8800}\\ \underline{85.58} / \slmT{50.45}\\ \slmT{\underline{90.93} / 88.92}\end{tabular} &
  \begin{tabular}[c]{@{}c@{}}\textbf{30.33} / \textbf{0.8983}\\ \textbf{87.80} / \slmT{\textbf{54.25}}\\ \slmT{\textbf{91.86} / \textbf{92.32}}\end{tabular} \\ \hline
\multirow{4}{*}{Denoising} &
  FUNSD$_{noise}$ &
  \multicolumn{1}{c|}{\begin{tabular}[c]{@{}c@{}}PSNR / SSIM $\uparrow$\\ Char. Acc. / \slmT{Word Acc.} (Paddle) (\%) $\uparrow$\\ \slmT{Char. Acc. / Word Acc. (Nougat) (\%) $\uparrow$}\end{tabular}} &
  \begin{tabular}[c]{@{}c@{}}39.92 / \underline{0.9955}\\ 96.30 / \slmT{79.79}\\ \slmT{87.70 / 83.45}\end{tabular} &
  \begin{tabular}[c]{@{}c@{}}35.04 / 0.9424\\ 95.09 / \slmT{76.02}\\ \slmT{87.17 / 84.33}\end{tabular} &
  \begin{tabular}[c]{@{}c@{}}\textbf{43.60} / \textbf{0.9982}\\ \textbf{97.58} / \slmT{\textbf{87.69}}\\ \slmT{\textbf{90.58} / \underline{86.60}}\end{tabular} &
  \begin{tabular}[c]{@{}c@{}}18.37 / 0.8606\\ 4.37 / \slmT{0.86}\\ \slmT{18.19 / 3.79}\end{tabular} &
  \begin{tabular}[c]{@{}c@{}}32.20 / 0.9837\\ 95.92 / \slmT{72.61}\\ \slmT{88.91 / 82.17}\end{tabular} &
  \begin{tabular}[c]{@{}c@{}}\underline{41.88} / \underline{0.9955}\\ \underline{96.73} / \slmT{\underline{83.53}}\\ \slmT{\underline{89.39} / \textbf{87.13}}\end{tabular} \\ \cline{2-9}
\multicolumn{1}{c|}{} &
  BCSD$_{noise}$ &
  \multicolumn{1}{c|}{\begin{tabular}[c]{@{}c@{}}PSNR / SSIM $\uparrow$\\ Char. Acc. / \slmT{Word Acc.} (Paddle) (\%) $\uparrow$\\ \slmT{Char. Acc. / Word Acc. (Nougat) (\%) $\uparrow$}\end{tabular}} &
  \begin{tabular}[c]{@{}c@{}}30.08 / 0.9320\\ 84.48 / \slmT{\textbf{51.50}}\\ \slmT{85.74 / 87.98}\end{tabular} &
  \begin{tabular}[c]{@{}c@{}}30.57 / 0.8396\\ 83.26 / \slmT{44.64}\\ \slmT{89.60 / 89.66}\end{tabular} &
  \begin{tabular}[c]{@{}c@{}}\underline{32.28} / \underline{0.9376}\\ \underline{86.37} / \slmT{\underline{50.95}}\\ \slmT{\underline{90.28} / 90.38}\end{tabular} &
  \begin{tabular}[c]{@{}c@{}}16.31 / 0.7267\\ 62.00 / \slmT{23.44}\\ \slmT{67.07 / 67.01}\end{tabular} &
  \begin{tabular}[c]{@{}c@{}}24.51 / 0.5752\\ 78.35 / \slmT{49.45}\\ \slmT{86.22 / \underline{90.79}}\end{tabular} &
  \begin{tabular}[c]{@{}c@{}}\textbf{35.15} / \textbf{0.9485}\\ \textbf{86.68} / \slmT{50.39}\\ \slmT{\textbf{91.75} / \textbf{91.94}}\end{tabular} \\ \hline
\multirow{4}{*}{CAR} &
  FUNSD$_{CAR}$ &
  \multicolumn{1}{c|}{\begin{tabular}[c]{@{}c@{}}PSNR / SSIM $\uparrow$\\ Char. Acc. / \slmT{Word Acc.} (Paddle) (\%) $\uparrow$\\ \slmT{Char. Acc. / Word Acc. (Nougat) (\%) $\uparrow$}\end{tabular}} &
  \begin{tabular}[c]{@{}c@{}}21.82 / 0.9484\\ 63.93 / \slmT{24.43}\\ \slmT{41.59 / 33.76}\end{tabular} &
  \begin{tabular}[c]{@{}c@{}}25.89 / 0.9665\\ 73.98 / \slmT{32.24}\\ \slmT{43.96 / 36.37}\end{tabular} &
  \begin{tabular}[c]{@{}c@{}}\textbf{28.10} / \underline{0.9745}\\ \underline{77.25} / \slmT{\textbf{47.52}}\\ \slmT{\textbf{61.33} / \underline{40.58}}\end{tabular} &
  \begin{tabular}[c]{@{}c@{}}18.39 / 0.8608\\ 3.39 / \slmT{0.73}\\ \slmT{16.51 / 3.28}\end{tabular} &
  \begin{tabular}[c]{@{}c@{}}26.62 / 0.9727\\ 73.54 / \slmT{31.89}\\ \slmT{45.18 / 36.48}\end{tabular} &
  \begin{tabular}[c]{@{}c@{}}\underline{27.44} / \textbf{0.9758}\\ \textbf{78.72} / \slmT{\underline{42.48}}\\ \slmT{\underline{57.96} / \textbf{48.24}}\end{tabular} \\ \cline{2-9}
\multicolumn{1}{c|}{} &
  BCSD$_{CAR}$ &
  \multicolumn{1}{c|}{\begin{tabular}[c]{@{}c@{}}PSNR / SSIM $\uparrow$\\ Char. Acc. / \slmT{Word Acc.} (Paddle) (\%) $\uparrow$\\ \slmT{Char. Acc. / Word Acc. (Nougat) (\%) $\uparrow$}\end{tabular}} &
  \begin{tabular}[c]{@{}c@{}}17.40 / 0.7297\\ 74.83 / \slmT{\underline{43.01}}\\ \slmT{82.51 / 82.44}\end{tabular} &
  \begin{tabular}[c]{@{}c@{}}22.62 / 0.7512\\ 77.46 / \slmT{42.27}\\ \slmT{87.39 / 87.42}\end{tabular} &
  \begin{tabular}[c]{@{}c@{}}\underline{24.85} / \underline{0.7851}\\ \underline{80.21} / \slmT{36.05}\\ \slmT{\underline{88.43} / 87.27}\end{tabular} &
  \begin{tabular}[c]{@{}c@{}}16.67 / 0.7189\\ 54.64 / \slmT{18.58}\\ \slmT{79.50 / 79.71}\end{tabular} &
  \begin{tabular}[c]{@{}c@{}}23.89 / 0.7694\\ 78.54 / \slmT{41.28}\\ \slmT{88.32 / \underline{88.33}}\end{tabular} &
  \begin{tabular}[c]{@{}c@{}}\textbf{25.59} / \textbf{0.7971}\\ \textbf{80.93} / \slmT{\textbf{46.69}}\\ \slmT{\textbf{88.59} / \textbf{88.59}}\end{tabular} \\ \hline
\multirow{4}{*}{Deshadowing} &
  OSR &
  \multicolumn{1}{c|}{\begin{tabular}[c]{@{}c@{}}PSNR / SSIM $\uparrow$\\ Char. Acc. / \slmT{Word Acc.} (Paddle) (\%) $\uparrow$\\ \slmT{Char. Acc. / Word Acc. (Nougat) (\%) $\uparrow$}\end{tabular}} &
  \begin{tabular}[c]{@{}c@{}}28.16 / 0.8944\\ 81.84 / \slmT{67.28}\\ \slmT{86.79 / \underline{87.52}}\end{tabular} &
  \begin{tabular}[c]{@{}c@{}}28.30 / 0.8544\\ 73.12 / \slmT{55.73}\\ \slmT{85.50 / 86.77}\end{tabular} &
  \begin{tabular}[c]{@{}c@{}}\textbf{28.66} / 0.9135\\ 81.33 / \slmT{68.45}\\ \slmT{\textbf{90.16} / 87.17}\end{tabular} &
  \begin{tabular}[c]{@{}c@{}}- / -\\ - / \slmT{-}\\ \slmT{- / -}\end{tabular} &
  \begin{tabular}[c]{@{}c@{}}\underline{28.53} / \underline{0.9304}\\ \underline{81.89} / \slmT{\textbf{72.92}}\\ \slmT{87.27 / 87.44}\end{tabular} &
  \begin{tabular}[c]{@{}c@{}}\textbf{28.66} / \textbf{0.9335}\\ \textbf{82.22} / \slmT{\underline{70.93}}\\ \slmT{\underline{87.73} / \textbf{88.35}}\end{tabular} \\ \cline{2-9}
\multicolumn{1}{c|}{} &
  Jung et al.'&
  \multicolumn{1}{c|}{\begin{tabular}[c]{@{}c@{}}PSNR / SSIM $\uparrow$\\ Char. Acc. / \slmT{Word Acc.} (Paddle) (\%) $\uparrow$\\ \slmT{Char. Acc. / Word Acc. (Nougat) (\%) $\uparrow$}\end{tabular}} &
  \begin{tabular}[c]{@{}c@{}}\underline{28.32} / 0.8921\\ 40.17 / \slmT{\underline{20.81}}\\ \slmT{35.87 / 37.57}\end{tabular} &
  \begin{tabular}[c]{@{}c@{}}27.69 / 0.8435\\ 37.60 / \slmT{19.42}\\ \slmT{32.23 / 21.92}\end{tabular} &
  \begin{tabular}[c]{@{}c@{}}27.96 / 0.9047\\ \underline{40.27} / \slmT{20.25}\\ \slmT{\underline{35.92} / 36.49}\end{tabular} &
  \begin{tabular}[c]{@{}c@{}}- / -\\ - / \slmT{-}\\ \slmT{- / -}\end{tabular} &
  \begin{tabular}[c]{@{}c@{}}28.25 / \underline{0.9082}\\ 40.04 / \slmT{\underline{20.81}}\\ \slmT{34.02 / \underline{37.83}}\end{tabular} &
  \begin{tabular}[c]{@{}c@{}}\textbf{28.35} / \textbf{0.9090}\\ \textbf{40.33} / \slmT{\textbf{20.95}}\\ \slmT{\textbf{36.73} / \textbf{38.06}}\end{tabular} \\ \hline
  &&\slmT{Average rank $\downarrow$} &
  \slmT{3.85} &
  \slmT{4.44} &
  \slmT{\underline{2.16}} &
  \slmT{5.94} &
  \slmT{3.20} &
  \slmT{\textbf{1.40}} \\ \hline
\end{tabular}
}
\end{table*}

\begin{figure*}[!t]
    \centering
    \includegraphics[width=\textwidth]{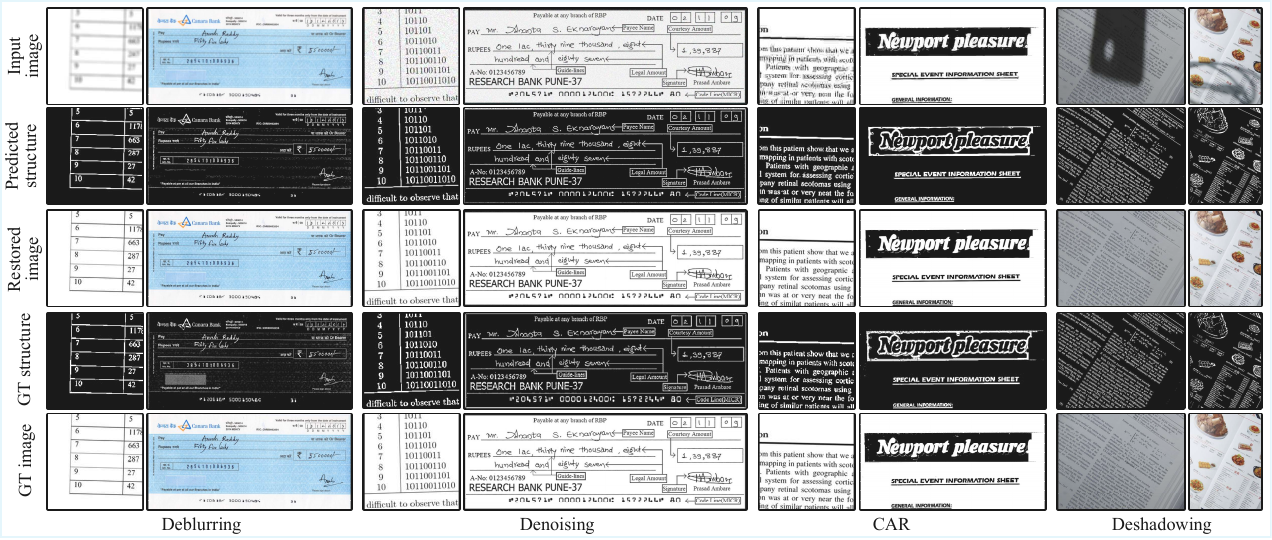} 
    \caption{Visual effects of our DocPure on diverse datasets with unseen styles (from top to bottom): degraded input images, our predicted structural maps, our restored images, GT structure maps, and GT images. The visual results 
    {show that our method recovers} 
     document content and structural details, 
     {suggesting} generalization capabilities across different document styles and degradation types.}
    \label{fig:our_all_tasks}
\end{figure*}

\subsection{Experimental results}

\subsubsection{Comparison with SOTA methods}

\slmT{Table~\ref{tab:1} presents a comprehensive quantitative comparison of DocPure against state-of-the-art methods across four restoration tasks. In the deblurring and CAR tasks, {our model achieves the best performance among the compared methods, attaining the highest PSNR and SSIM scores, as well as the highest character- and word-level accuracies evaluated using both PaddleOCR and Nougat.} In the denoising task, our model exhibits the second-best performance for the {six} metrics. Regarding the real-world deshadowing task, while DocRes shows a marginal advantage in PSNR {and Nougat word accuracy}, our model surpasses it in structural fidelity and text recognizability, achieving the highest values of SSIM (0.9532), PaddleOCR character accuracy (65.69\%), {PaddleOCR word accuracy (56.23\%)}, and Nougat character accuracy (87.85\%).}

{Fig.~\ref{compare1} provides a visual comparison of our method with existing approaches.} As shown in the deblurring task (first row), DocPure reconstructs text with {clearer}  structures and continuous strokes compared to other methods. In the denoising scenario (second row), our method robustly recovers clean details from Gaussian noise corruption, avoiding the over-smoothing or residual noise commonly observed in baselines. For the CAR task (third row), the model 
{handles} complex high-frequency compression noise, 
{reducing} blocking artifacts while maintaining the integrity of text strokes. Qualitative results for deshadowing (fourth row) demonstrate that DocPure restores uniform illumination and maintains stable font morphology, 
{mitigating} the background residues and semantic blurring prevalent in existing all-in-one models. {Since DocNLC requires cross-task pairing for the training data, it cannot be trained with the unpaired deshadowing dataset.}

{The consistently strong performance across these diverse tasks suggests the effectiveness of our unified framework, rather than relying on task-specific engineering.
} The successful recovery of fine-grained structure details {supports} the effectiveness of DASAE in {predicting} clear structural priors from corrupted inputs. The robust performance across multiple tasks underscores the efficacy of leveraging the distinctiveness of degradations in the wavelet domain and endowing the model with degradation awareness through wavelet-spatial interaction.

\subsubsection{Generalization ability assessment}
To validate the model's robustness, we evaluated the cross-domain generalization capability of DocPure on document styles and layouts that were unseen during training.
\slmT{As shown in Table~\ref{tab:compare_add}, our method consistently maintains top-tier performance across {unseen} datasets.
For example, in the deshadowing task, DocPure achieves the best performance on the OSR dataset in terms of PSNR (28.66 dB), SSIM (0.9335), and the PaddleOCR character accuracy (82.22\%), and the second-best Nougat character accuracy (87.73\%); On the {real-world} Jung et al. datasets, DocPure further achieves
 {the best performance among the compared methods} for all {six} evaluation metrics, highlighting the effectiveness of our unified architecture.
DocPure's {performance} on challenging unseen {document styles and layouts} attests to the model's strong feature adaptability.}

Fig.~\ref{fig:our_all_tasks} visually demonstrates the model's robustness against unseen styles. Even when handling document layouts with complex fonts or variable spacing, our method reconstructs characters with sharp boundaries and continuous strokes, avoiding the fractured text artifacts common in other methods. In contrast to some compared methods that suffer from semantic blurring, DocPure maintains morphological consistency, validating its degradation-aware generalization.

\begin{table*}[t]
\caption{Comparison of model properties among different document restoration methods. Our DocPure is the only 
{model among the compared methods}
supporting all three properties: all-in-one inference, prompt-free \slmT{inference}, and cross-task pairing-free training data.}
\centering
\label{tab:method_comparison}

{%
\setlength{\tabcolsep}{10pt}
\begin{tabular}{c|ccccc|c}
\hline
Model property &
  DE-GAN~\cite{souibgui2020gan} &
  DocDiff~\cite{yang2023docdiff} &
  DocStain~\cite{li2025high} &
  DocNLC~\cite{wang2024docnlc} &
  DocRes~\cite{zhang2024docres} &
  DocPure (ours) \\ \hline
All-in-one inference & 
  \xmark & \xmark & \xmark & \checkmark & \checkmark & \checkmark \\ 
Prompt-free \slmT{inference} & 
  \xmark & \xmark & \xmark & \checkmark & \xmark & \checkmark \\ 
Cross-task pairing-free & 
  \checkmark & \checkmark & \checkmark & \xmark & \checkmark & \checkmark \\ \hline
\end{tabular}%
}
\end{table*}

\begin{table}[t]
\caption{Statistics of number of parameters (\#Params) and computational complexity (GFLOPs) \slmT{at the resolution of $256 \times 256$.}}
\centering
\label{tab:params_flops}
 {%
 \begin{tabular}{c|ccc}
 \hline
 {Metric} &
   DE-GAN~\cite{souibgui2020gan} &
   DocDiff~\cite{yang2023docdiff} &
   DocStain~\cite{li2025high} \\ \hline
 \#Params (M) &
   30.00 &
   8.20 &
   6.86 \\ 
  GFLOPs  &
   109.00&
   4876.50 &
   283.90 \\ \hline
 {Metric} &
   DocNLC~\cite{wang2024docnlc} &
   DocRes~\cite{zhang2024docres} &
   DocPure (ours) \\ \hline
 \#Params (M) &
   7.80 &
   15.20 &
   38.15 \\ 
 GFLOPs &
   34.80 &
   183.00 &
   446.82 \\ \hline
 \end{tabular}%
 }
 \end{table}

\subsubsection{Discussion on model properties}

As illustrated in Table~\ref{tab:method_comparison}, existing document image restoration methods struggle to simultaneously satisfy the requirements of unified modeling, prompt-free inference, and {cross-task pairing-free training data}. \slmT{First of all,} previous general-task document restoration models, such as DE-GAN, DocDiff, and DocStain, require training separate weights for different degradation types (\ie several models for several degradations), which scales poorly in real-world scenarios. \slmT{On the contrary, our unified framework enables adaptive high-quality document restoration by taking advantage of the proposed degradation-informed routing regularization mechanism.}
\slmT{Second, }DocRes achieves unified modeling but relies heavily on explicit task prompts at inference to specify the degradation category. Consequently, it fails to perform degradation-aware restoration when the degradation prior is unknown at inference. 
In contrast, our DocPure eliminates this dependency \slmT{by leveraging the proposed DASAE to predict the structure map to guide the prompt-free degradation-aware inference}.
\slmT{Third,} although DocNLC achieves prompt-free unified restoration, its contrastive learning framework imposes a stringent constraint on dataset construction. Specifically, it requires strictly paired multi-degradation data (\ie multiple distinct degradation types must correspond to the exact same clean ground truth), which is highly restrictive and costly to synthesize or collect at scale. \slmT{Fortunately,} our DocPure overcomes this limitation by operating effectively without such rigorous pairing demands. In conclusion, DocPure stands out as the only method that achieves prompt-free document restoration with a single unified model \slmT{while {avoiding} the strict data pairing constraint. This makes our approach {more} practical for diverse
{document restoration settings}}.

\begin{table}[t]
\slmT{
\centering
\caption{Statistics of efficiency under different number of sub-encoders $k$ at the resolution of $1024 \times 1024$.}
\label{tab:model_comparison_merged}
\setlength{\tabcolsep}{1pt}
\resizebox{\linewidth}{!}
{
\begin{tabular}{lcccccc}
\toprule
\multirow{2}{*}{$k$ value} & \multicolumn{3}{c}{{DASAE}} & \multicolumn{3}{c}{{Full model (DocPure)}} \\
\cmidrule(lr){2-4} \cmidrule(lr){5-7} 
 & {Params} & {Latency} & {GPU memory} & {Params} & {Latency} & {GPU memory} \\
\midrule
$k=1$ & 4.06M & 55.5ms  & 2.194GB & 34.60M & 187.0ms & 4.929GB \\
$k=2$ & 5.23M & 78.7ms  & 2.209GB & 35.81M & 209.6ms & 4.936GB \\
$k=3$ & 6.40M & 100.0ms & 2.491GB & 36.98M & 231.9ms & 4.942GB \\
$k=4$ & 7.57M & 122.4ms & 3.014GB & 38.15M & 253.9ms & 4.948GB \\
\bottomrule
\end{tabular}
}
}
\end{table}

\begin{table*}[t]
\centering
\caption{Ablation study of the proposed components on various document restoration tasks. Our DocPure achieves the best overall performance. The best performance is highlighted in \textbf{bold}.}
\label{tab:ablation_esen_werm}
{
\begin{tabular}{c|cc|cc|cc|cc|cc}
\hline
\multirow{2}{*}{\makebox[0.15\textwidth][c]{Configuration}} & \multirow{2}{*}{\makebox[0.07\textwidth][c]{DASAE}} & \multirow{2}{*}{\makebox[0.07\textwidth][c]{SWIM}} & 
\multicolumn{2}{c|}{\makebox[0.134\textwidth][c]{Deblurring}} & 
\multicolumn{2}{c|}{\makebox[0.134\textwidth][c]{Denoising}} & 
\multicolumn{2}{c|}{\makebox[0.134\textwidth][c]{CAR}} & 
\multicolumn{2}{c}{\makebox[0.134\textwidth][c]{Deshadowing}} \\ 
\cline{4-11}
 & & & PSNR$\uparrow$ & SSIM$\uparrow$ & PSNR$\uparrow$ & SSIM$\uparrow$ & PSNR$\uparrow$ & SSIM$\uparrow$ & PSNR$\uparrow$ & SSIM$\uparrow$ \\ 
\hline
Baseline & \xmark & \xmark & 28.80 & 0.9838 & 40.13 & 0.9960 & 29.68 & 0.9850 & 32.01 & 0.9513 \\
w/o DASAE & \xmark & \cmark & 31.66 & 0.9901 & 40.70 & 0.9968 & 30.88 & 0.9885 & 32.64 & 0.9521 \\
w/o SWIM & \cmark & \xmark & 29.84 & 0.9861 & 40.15 & 0.9960 & 30.04 & 0.9861 & 32.22 & 0.9517 \\
Ours & \cmark & \cmark & \textbf{32.55} & \textbf{0.9912} & \textbf{41.26} & \textbf{0.9973} & \textbf{31.34} & \textbf{0.9893} & \textbf{33.31} & \textbf{0.9532} \\ 
\hline
\end{tabular}
}
\end{table*}

\subsubsection{Analysis on computational complexity}

 {Table~\ref{tab:params_flops} provides statistics of the number of parameters and computational complexity (GFLOPs). From the table, we can see that DocDiff suffers from massive computational overhead (4876.50 GFLOPs) due to its iterative generation process. DocNLC has fewer parameters and fewer computations, but its restoration quality in terms of PSNR and downstream character accuracy falls considerably across multiple degradation scenarios.
 \rev{While DocPure is substantially lighter than the diffusion-based DocDiff (446.82 GFLOPs vs. 4876.50 GFLOPs), it is heavier than other non-diffusion all-in-one baselines such as (34.80 GFLOPs) and DocRes (183.00 GFLOPs). This reflects that the design choice of DocPure prioritizes restoration quality and degradation awareness over computational minimalism.}

\slmT{Table~\ref{tab:model_comparison_merged} reports statistics of the number of parameters, inference latency, and peak GPU memory usage for both DASAE and the full model under different numbers of sub-encoders $k$ at the resolution of $1024 \times 1024$. With the increase of $k$, the total computational overhead of our multi-branch DocPure increases moderately (around 20ms) compared to the inference latency (187ms) of the single-branch model ($k=1$). The structure predictor is required for every degraded input. The results exhibit a good trade-off between the performance improvement and the computational overhead achieved by our multi-branch sub-encoder architecture.} \rev{Our multi-branch design is scalable. A sub-encoder can be added to our DASAE for a new degradation type. This requires training the new sub-encoder, incrementally fine-tuning DASAE, and adapting the restoration backbone with the updated structure predictor. Based on the one-to-one correspondence, the number of sub-encoders grows linearly with the number of degradation types.} As shown in Table~\ref{tab:model_comparison_merged}, the number of parameters of the full models increases slightly with nearly stable peak GPU memory usage.

\begin{figure}[t]
    \centering
    \includegraphics[width=\columnwidth]{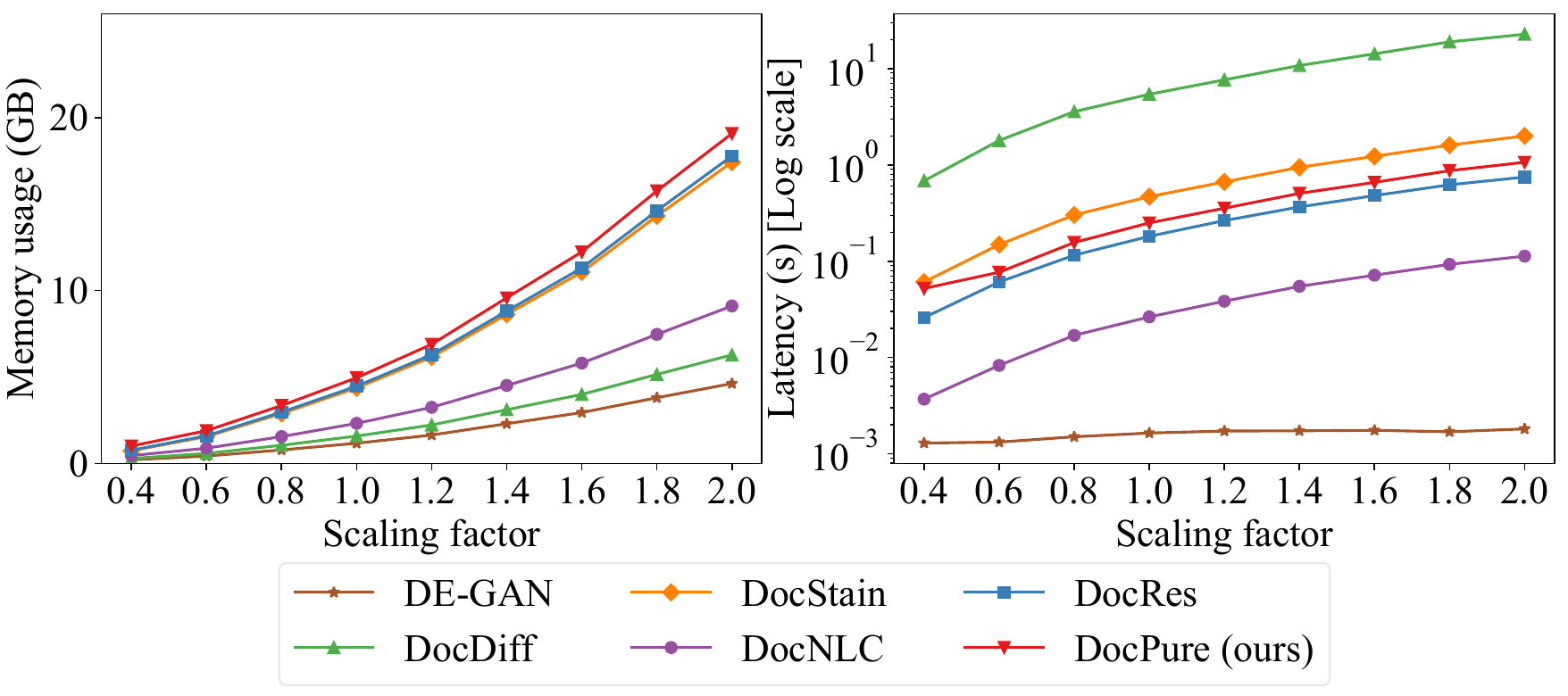}
    \caption{\slmT{Peak GPU memory usage (left) and average inference latency (right) of compared methods at various scaling factors applied to a $1024 \times 1024$ resolution image.}}
    \label{mem_usage}
\end{figure}

\slmT{In Fig.~\ref{mem_usage}, we visualize the maximum GPU memory usage and inference latency for the compared models {when processing images of various resolutions}. Our method exhibits slightly higher memory usage compared to the DocRes and DocStain baselines. Although the multi-branch architecture suggests higher computational costs, we can see that our method still maintains a moderate computational overhead. For example, at a resolution of $1024 \times 1024$, the inference latency of our DocPure is 253.9ms per image. While this is slower than DocRes (181.3ms), it is significantly faster than DocDiff (5413.9ms) and DocStain (467.6ms).}

\subsection{Ablation study}

Table~\ref{tab:ablation_esen_werm} quantitatively validates the contributions of \rev{the predicted structure map and the wavelet interaction with the proposed DASAE and SWIM modules, respectively}. First, the w/o DASAE variant fails to achieve the optimal performance. This performance drop confirms that without the explicit structural priors learned by the 
\slmT{DASAE}, the restoration network lacks precise structure guidance, leading to suboptimal stroke reconstruction. Second, we observe a distinct performance degradation in the w/o SWIM configuration. This quantitative gap provides evidence that relying solely on spatial convolutions is 
{less effective for handling} complex frequency-dependent degradations. The wavelet-domain interaction is {important} for recovering high-frequency structure details. The {best overall performance} of the full DocPure framework demonstrates the synergistic effect of combining degradation-aware structural extraction with frequency-guided modulation.

\begin{figure}[t]
    \centering
    \includegraphics[width=\columnwidth]{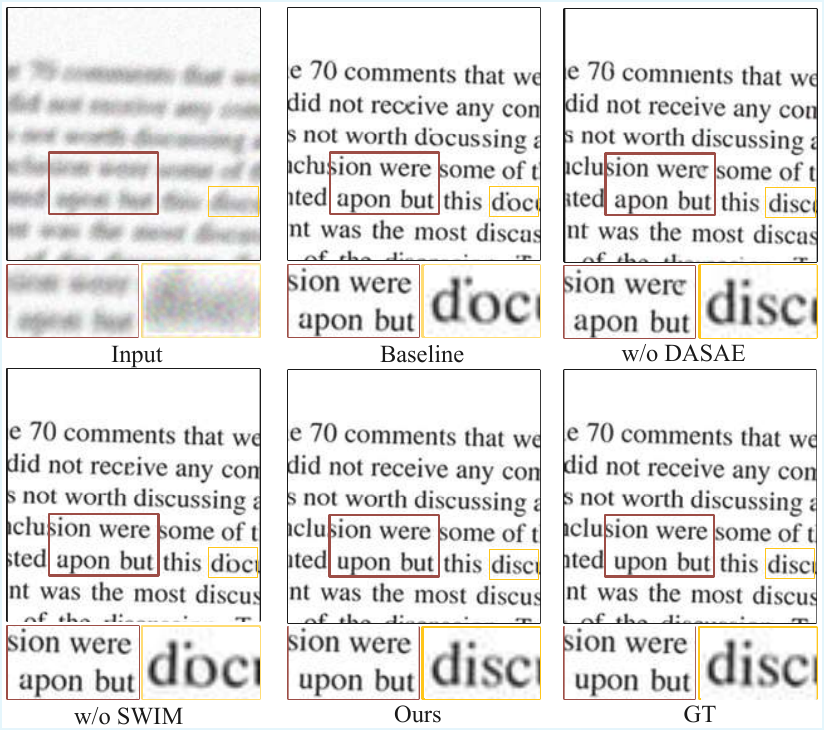}
    \caption{Visual ablation comparison of different components. The baseline and variants without specific modules (w/o DASAE, w/o SWIM) fail to recover semantically correct text, resulting in character substitution errors. In contrast, our full model integrates all proposed modules to restore sharp, legible, and semantically accurate text, matching the GT. }
    \label{fig:ab_compare}
\end{figure}

\begin{table*}[t]
\centering
\slmT{
\caption{{Sensitivity analysis} of loss weights $\lambda_{bce}$, $\lambda_{dice}$, and $\lambda_{bd}$. 
For each parameter group, only the corresponding weight is varied, while the other two are fixed to the default configuration 
$(\lambda_{bce}, \lambda_{dice}, \lambda_{bd})=(0.30,0.50,0.35)$. 
The best performance in each parameter group is highlighted in \textbf{bold}.}
\label{tab:ablation_weights}
\scriptsize
\setlength{\tabcolsep}{2.2pt}
\renewcommand{\arraystretch}{0.92}
\resizebox{\textwidth}{!}{
\begin{tabular}{cc|ccc|ccc|ccc|ccc}
\hline
\multirow{2}{*}{Parameter} & \multirow{2}{*}{Value} &
\multicolumn{3}{c|}{Deblurring} &
\multicolumn{3}{c|}{Denoising} &
\multicolumn{3}{c|}{CAR} &
\multicolumn{3}{c}{Deshadowing} \\
\cline{3-14}
 & & 
PSNR$\uparrow$ & SSIM$\uparrow$ & Char. Acc.$\uparrow$ &
PSNR$\uparrow$ & SSIM$\uparrow$ & Char. Acc.$\uparrow$ &
PSNR$\uparrow$ & SSIM$\uparrow$ & Char. Acc.$\uparrow$ &
PSNR$\uparrow$ & SSIM$\uparrow$ & Char. Acc.$\uparrow$ \\
\hline

\multirow{5}{*}{$\lambda_{bce}$}
& 0.00 & 32.44 & 0.9908 & 55.84 & 41.20 & 0.9971 & 72.29 & 31.23 & 0.9890 & 53.09 & 33.22 & 0.9526 & 87.31 \\
& 0.15 & 32.48 & 0.9911 & 56.13 & \textbf{41.27} & \textbf{0.9973} & \textbf{72.33} & 31.32 & 0.9891 & 53.29 & 33.30 & 0.9532 & 87.76 \\
& 0.30 & 32.55 & \textbf{0.9912} & 56.31 & 41.26 & \textbf{0.9973} & 72.30 & \textbf{31.34} & \textbf{0.9893} & \textbf{53.40} & \textbf{33.31} & 0.9532 & \textbf{87.85} \\
& 0.45 & \textbf{32.56} & \textbf{0.9912} & \textbf{56.34} & 41.24 & \textbf{0.9973} & 72.23 & 31.26 & 0.9891 & 53.15 & \textbf{33.31} & \textbf{0.9533} & 87.84 \\
& 0.60 & 32.52 & 0.9910 & 55.93 & 41.24 & \textbf{0.9973} & 72.22 & 31.27 & 0.9891 & 53.13 & 33.29 & 0.9531 & 87.73 \\
\hline

\multirow{5}{*}{$\lambda_{dice}$}
& 0.00 & 32.42 & 0.9909 & 55.78 & 41.22 & \textbf{0.9973} & 72.22 & 31.21 & 0.9888 & 52.97 & 33.16 & 0.9524 & 87.04 \\
& 0.25 & 32.50 & 0.9911 & 56.07 & 41.24 & \textbf{0.9973} & 72.26 & 31.28 & 0.9891 & 53.22 & 33.28 & 0.9530 & 87.77 \\
& 0.50 & 32.55 & 0.9912 & 56.31 & \textbf{41.26} & \textbf{0.9973} & 72.30 & 31.34 & \textbf{0.9893} & \textbf{53.40} & \textbf{33.31} & \textbf{0.9532} & \textbf{87.85} \\
& 0.75 & \textbf{32.59} & \textbf{0.9913} & \textbf{56.42} & \textbf{41.26} & \textbf{0.9973} & \textbf{72.33} & \textbf{31.36} & \textbf{0.9893} & 53.37 & 33.27 & 0.9528 & 87.82 \\
& 1.00 & 32.49 & 0.9910 & 56.06 & 41.25 & \textbf{0.9973} & 72.29 & 31.31 & 0.9892 & 53.28 & 33.28 & 0.9527 & 87.66 \\
\hline

\multirow{5}{*}{$\lambda_{bd}$}
& 0.00 & 32.45 & 0.9909 & 55.88 & 41.24 & \textbf{0.9973} & 72.26 & 31.25 & 0.9891 & 53.12 & 33.22 & 0.9527 & 87.33 \\
& 0.10 & 32.47 & 0.9910 & 55.90 & 41.24 & \textbf{0.9973} & 72.23 & 31.27 & 0.9891 & 53.14 & 33.24 & 0.9527 & 87.36 \\
& 0.35 & \textbf{32.55} & \textbf{0.9912} & \textbf{56.31} & \textbf{41.26} & \textbf{0.9973} & \textbf{72.30} & \textbf{31.34} & \textbf{0.9893} & \textbf{53.40} & 33.31 & \textbf{0.9532} & 87.85 \\
& 0.50 & \textbf{32.55} & 0.9911 & 56.29 & \textbf{41.26} & \textbf{0.9973} & 72.28 & 31.28 & 0.9892 & 53.25 & \textbf{33.33} & \textbf{0.9532} & \textbf{87.91} \\
& 0.70 & 32.51 & 0.9910 & 56.03 & 41.25 & \textbf{0.9973} & 72.27 & 31.27 & 0.9892 & 53.22 & 33.31 & \textbf{0.9532} & 87.84 \\
\hline
\end{tabular}
}}
\end{table*}

Fig.~\ref{fig:ab_compare} illustrates the visual results in the absence of specific modules. In the absence of structure maps predicted by DASAE serving as internal priors, the restoration network loses structural constraints. Without such structure maps, the model struggles to preserve character strokes and capture complex text structures. Upon the removal of the SWIM module, the model {loses degradation-aware spectral guidance}. In this scenario, the model tends to project multiple heterogeneous degradation patterns into a unified feature space for processing, leading to severe cross-task interference. Due to the lack of targeted guidance and enhancement for high-frequency components, the restored text structures are {not sufficiently sharp and may also exhibit} noticeable ghosting or artifacts.

\slmT{We conducted {a sensitivity analysis of} the loss weights $\lambda_{bce}$, $\lambda_{dice}$, and $\lambda_{bd}$. As reported in Table~\ref{tab:ablation_weights}, the overall performance varies slightly across different weight settings, showing that our DocPure is not overly sensitive to a particular weight choice. The default setting ($\lambda_{bce}$, $\lambda_{dice}$, $\lambda_{bd}$) = (0.30, 0.50, 0.35) achieves the best overall performance across the four document restoration tasks. The BCE loss provides stable pixel-level supervision {for dense structure-map prediction}, the Dice loss encourages region-level consistency between text foreground and background, and the boundary loss improves stroke contour localization.}

\begin{table}[t]
\centering
\slmT{
\caption{Comparison of utilizing different wavelet bases for frequency-domain decomposition.}
\label{tab:wavelet_ablation}
\footnotesize 
\setlength{\tabcolsep}{2pt} 
\resizebox{\columnwidth}{!}{
\begin{tabular}{c|cc|cc|cc|cc}
\hline
\multirow{2}{*}{Wavelet basis} & 
\multicolumn{2}{c|}{Deblurring} & 
\multicolumn{2}{c|}{Denoising} & 
\multicolumn{2}{c|}{CAR} & 
\multicolumn{2}{c}{Deshadowing} \\ 
\cline{2-9}
& PSNR$\uparrow$ & SSIM$\uparrow$ & PSNR$\uparrow$ & SSIM$\uparrow$ & PSNR$\uparrow$ & SSIM$\uparrow$ & PSNR$\uparrow$ & SSIM$\uparrow$ \\ 
\hline
Haar~\cite{mallat1989wavelet} & \textbf{32.55} & \textbf{0.9912} & \textbf{41.26} & \textbf{0.9973} & 31.34 & \textbf{0.9893} & 33.31 & 0.9532 \\ 
Db2~\cite{daubechies1988orthonormal}& 32.39 & 0.9898 & 41.21 & 0.9966 & \textbf{31.51} & 0.9889 & \textbf{33.46} & \textbf{0.9539} \\
\hline
\end{tabular}
}}
\end{table}

\begin{table}[t]
\centering
\slmT{
\caption{
Ablation study on the design choices of DASAE and SWIM with simpler alternatives:
(A) a single shared encoder with task-label auxiliary loss replacing the multi-branch \slmT{degradation} encoder in DASAE;
(B) standard wavelet concatenation without cross-attention;
(C) Sobel edge mask priors instead of learned structure maps;
(D) FiLM modulation without the full CFAM design;
(E) a degradation classifier plus conditional normalization.
}
\label{tab:arch_ablation}
\footnotesize
\setlength{\tabcolsep}{3pt}
\resizebox{\columnwidth}{!}{
\begin{tabular}{c|cc|cc|cc|cc}
\hline
\multirow{2}{*}{Setting} & 
\multicolumn{2}{c|}{Deblurring} & 
\multicolumn{2}{c|}{Denoising} & 
\multicolumn{2}{c|}{CAR} & 
\multicolumn{2}{c}{Deshadowing} \\ 
\cline{2-9}
& PSNR$\uparrow$ & SSIM$\uparrow$ & PSNR$\uparrow$ & SSIM$\uparrow$ & PSNR$\uparrow$ & SSIM$\uparrow$ & PSNR$\uparrow$ & SSIM$\uparrow$ \\ 
\hline
A & 32.38 & 0.9905 & 41.22 & 0.9972 & 31.27 & 0.9890 & 33.06 & 0.9522 \\
B & 30.52 & 0.9865 & 40.58 & 0.9956 & 30.13 & 0.9864 & 32.33 & 0.9521 \\
C & 31.21 & 0.9891 & 41.17 & 0.9965 & 31.18 & 0.9871 & 32.72 & 0.9516 \\
D & 31.97 & 0.9905 & 40.94 & 0.9966 & 31.20 & 0.9868 & 32.69 & 0.9521 \\
E & 30.93 & 0.9887 & 40.59 & 0.9962 & 30.62 & 0.9878 & 32.42 & 0.9528 \\
Ours& \textbf{32.55} & \textbf{0.9912} & \textbf{41.26} & \textbf{0.9973} & \textbf{31.34} & \textbf{0.9893} & \textbf{33.31} & \textbf{0.9532} \\
\hline
\end{tabular}
}}
\end{table}

\slmT{Table~\ref{tab:wavelet_ablation} provides the comparison of utilizing the Haar wavelet~\cite{mallat1989wavelet} and the Daubechies 2 (db2) wavelet~\cite{daubechies1988orthonormal} for frequency-domain decomposition within SWIM. The Haar wavelet is 
suitable for preserving abrupt edges in sharp font strokes, while the db2 wavelet is more suitable for modeling the restoration of low-frequency continuous signals {such as} shadow gradients. The experimental results show that the performance of the two choices is comparable overall. Haar achieves better performance in deblurring and denoising, while db2 {performs better} in deshadowing.}

\slmT{As shown in Table~\ref{tab:arch_ablation}, we conducted an ablation study to justify the design choices of DASAE and SWIM by comparing them with {several baseline variants}.
The tested alternatives include: (A) a single shared encoder with task-label auxiliary loss, (B) standard wavelet concatenation without cross-attention, (C) Sobel mask priors instead of learned structure maps, (D) FiLM modulation without the full CFAM design, and (E) a degradation classifier plus conditional normalization. The results demonstrate that the design choices of DASAE and SWIM consistently {achieve} the best performance across all four restoration tasks. {Therefore, this ablation study supports the proposed design choices of degradation-aware structure prediction and structure-guided frequency modulation for unified document restoration.}}

\begin{figure}[t]
    \centering
    \includegraphics[width=\columnwidth]{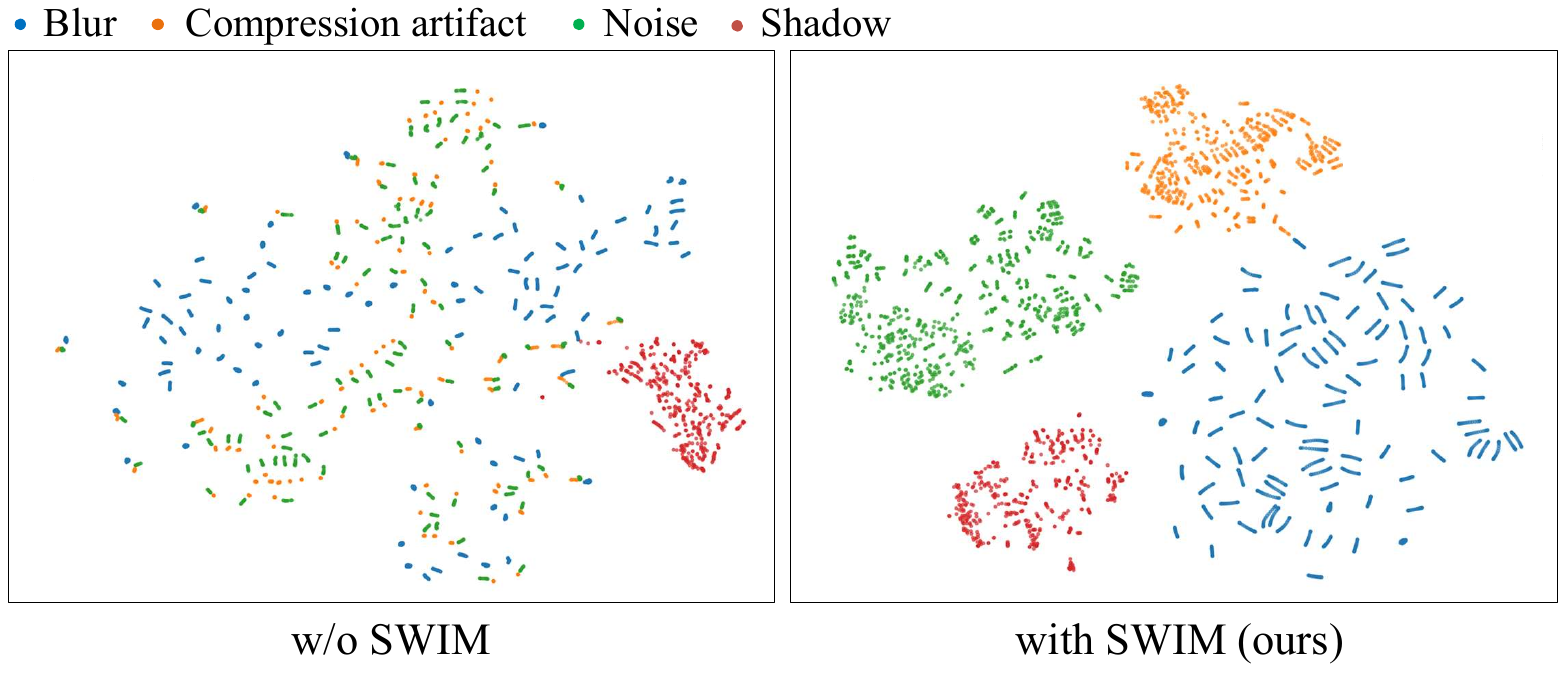}
    \caption{{Comparison of t-SNE visualization of latent feature distributions extracted from the model bottlenecks without and with the SWIM block.} The inclusion of SWIM results in well-separated clusters, indicating improved degradation discriminability.}
    \label{wermcompare}
\end{figure}

{Fig.~\ref{wermcompare} visualizes the comparison of latent feature distributions extracted from the models with and without SWIM.}  \slmT{Compared to the baseline without SWIM, our model successfully maps different degradations to well-separated clusters. This visualization 
provides qualitative evidence that SWIM improves degradation-discriminative representations, supporting input-adaptive restoration without manual task prompts}.

\begin{figure}[t]
    \centering
    \includegraphics[width=\columnwidth]{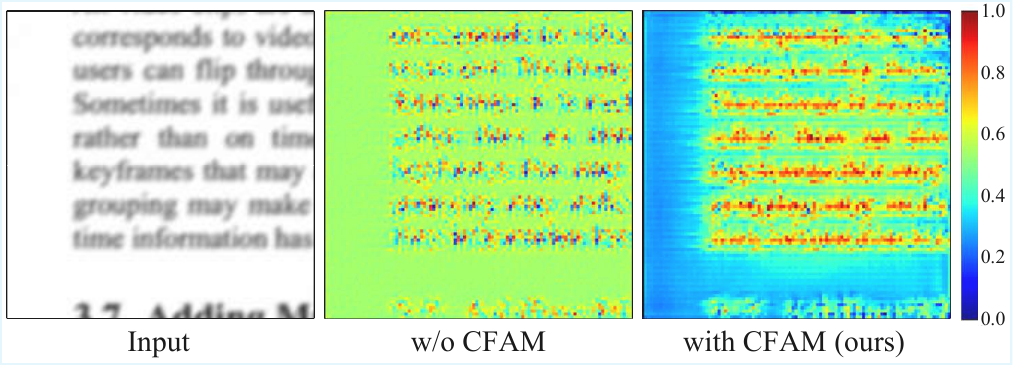}
    \caption{Feature heatmaps without and with the CFAM module. The lack of the CFAM module hinders the separation of text from background, causing ambiguous feature activations. Conversely, CFAM {enhances} localization capability, guiding the model to focus on informative text features.}
    \label{WLFM}
\end{figure}

To visually demonstrate the effectiveness of CFAM, Fig.~\ref{WLFM} presents the feature heatmaps with and without the integration of CFAM. In the configuration without CFAM, the feature activation distribution appears relatively diffuse. Consequently, the model struggles to effectively differentiate between text strokes and background noise, leading to severe ambiguity in feature responses. In contrast, upon introducing CFAM, the low-frequency structural priors {help} calibrate the high-frequency feature distribution through the dual mechanism of spatial gating and channel modulation. This enables the model to focus more precisely on informative text regions, enhancing the saliency of text structures while suppressing artifacts in non-text regions.

\begin{figure}[!t]
    \centering
    \slmT{
    \includegraphics[width=\columnwidth]{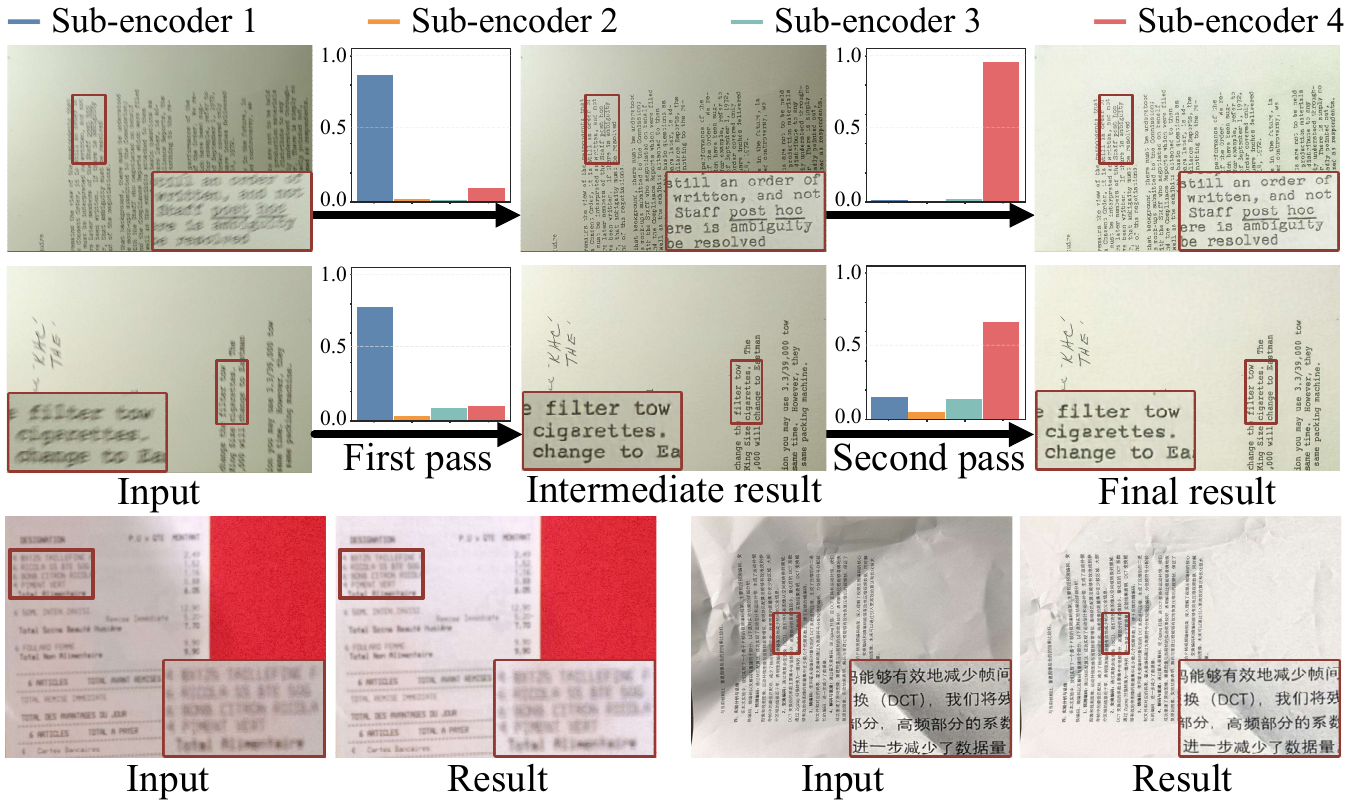}
    \caption{Examples of our restoration results under real-world mixed degradation scenarios. Our method can mitigate real-world degradations {in} documents with moderate blur and shadow (top), while tending to fail to restore extremely blurred small fonts or paper-crease degradations unseen during training (bottom).}
    \label{fig:failure&mix}
    }
\end{figure}

\slmT{\subsection{Discussion on real-world document restoration}}

In Table~\ref{tab:1} and Table~\ref{tab:compare_add}, we have demonstrated that our method is capable of restoring documents on a single real-world degradation with the real-world deshadowing datasets (RDD~\cite{zhang2023RDD} and Jung et al.'s dataset~\cite{jung2018jun}). Here we further test the performance of our method under real-world mixed degradation scenarios, \rev{where multiple degradation types coexist in a single image}. Since our multi-branch model design focuses on the restoration of a single dominant degradation, we ran multiple inference passes on mixed degradations to obtain the restored result. As shown in Fig.~\ref{fig:failure&mix} (top), our method can mitigate real-world degradations in documents  with moderate blur and shadow. \rev{The routing mechanism produces a weighted combination of sub-encoder features, where the dominant degradation is distinguished by the largest weight. The figure shows that our method successfully performs deblurring in the first inference pass, followed by deshadowing in the second.} However, our method tends to fail to restore extremely blurred small fonts or paper-crease degradations unseen during training. The failure cases in Fig.~\ref{fig:failure&mix} (bottom) show that the restored results still contain incomplete text details or local residual degradations. This is mainly because our method is currently designed around a known set of trained degradations. Nevertheless, this experiment indicates that our method has potential for real-world mixed degradation scenarios.}

\section{Conclusions}

We propose DocPure, a prompt-free degradation-aware framework for unified document image restoration.
By explicitly identifying degradation types to guide structural priors via \slmT{degradation-informed routing regularization} in the latent space, while implicitly perceiving degradation patterns through collaborative interaction in the wavelet domain, our DocPure achieves an effective balance between the restoration accuracy and generalization capability {across diverse document types}. \rev{The degradation-label supervision is only applied during training, and no manual task prompts or degradation labels are required at inference.}
The proposed degradation-aware structure auto-encoder and structure-guided wavelet interaction module are integrated into a unified U-shaped architecture. This design enables the model to identify diverse degradation modes via spectral signatures, while leveraging cross-frequency adaptive modulation to guide features toward structure-sharp representations and suppress high-frequency artifacts. Extensive experiments on 12 datasets across four representative document restoration tasks demonstrate that our {prompt-free} DocPure not only achieves
{strong performance compared with SOTA methods}
 without requiring task prompts, but also exhibits robustness on unseen document styles. 

\slmT{\textbf{Limitations and future work.} Our method comes with certain limitations. First, as mentioned beforehand, our method is based on a known set of trained degradations. Adapting to a new degradation {category} may require additional incremental training or an expansion of the routing-space capacity. \rev{Second, for mixed degradations, spatially non-uniform degradations, or degradations outside the training degradation set, the model does not have a dedicated routing pathway, and the weighted combination of sub-encoder features may not fully capture the complex degradation distribution.} Developing a more scalable blind document restoration framework remains an interesting future direction. Third, the memory consumption becomes high when restoring high-resolution documents, which limits the deployment on resource-limited edge devices. 
We will investigate a more lightweight model that utilizes techniques such as efficient context modeling~\cite{EchoSR} and model distillation.
In addition, we would like to extend our method to general image restoration applications in the near future.}


\clearpage

\begin{center}
    \textbf{\large SUPPLEMENTARY DOCUMENT}
\end{center}

\setcounter{section}{0}
\section{More implementation details}
\label{sec:training}

\begin{algorithm}[!thpb] 
    \caption{Pseudocode of DocPure}
    \label{alg:docpure_main}
    \begin{algorithmic}[1]
    \State Predict structure map $\mathbf{I}_M \leftarrow \operatorname{DASAE}(\mathbf{I})$ \hfill  
    \State Extract shallow features $\mathbf{Y}_0 \leftarrow \operatorname{Conv}(\mathbf{I})$ \hfill 
        \For{$l = 1$ to $3$}  \hfill \textcolor{gray}{\# Encoding phase}
            \State $\mathbf{Y}'_l \leftarrow \operatorname{TransformerBlocks}(\mathbf{Y}_{l-1})$ 
            \State $\mathbf{E}_l \leftarrow \mathbf{Y}'_l$ \hfill \textcolor{gray}{\# Save for skip connection}
            \State $\mathbf{Y}_l \leftarrow \operatorname{PixelUnshuffle}(\operatorname{Conv}(\mathbf{Y}'_l))$ \hfill \textcolor{gray}{\# Downsample}
        \EndFor
        \State $\mathbf{Y}_4 \leftarrow \operatorname{TransformerBlocks}(\mathbf{Y}_3)$ \hfill \textcolor{gray}{\# Bottleneck layer}
        
        \For{$l = 3$ down to $1$}\hfill\textcolor{gray}{\# Decoding phase}
            \State $\mathbf{Y}_{up} \leftarrow \operatorname{PixelShuffle}(\operatorname{Conv}(\mathbf{Y}_{l+1}))$ \hfill \textcolor{gray}{\# Upsample}
            \State $\mathbf{Y}' \leftarrow \operatorname{SWIM}(\mathbf{Y}_{up}, \mathbf{I}, \mathbf{I}_M)$ 
            \State $\mathbf{Y}'' \leftarrow [\mathbf{Y}', \mathbf{E}_l]$ \hfill \textcolor{gray}{\# Skip connection}
            \If{$l > 1$}
                \State $\mathbf{Y}_l \leftarrow \operatorname{TransformerBlocks}(\mathbf{Y}'')$
            \EndIf
        \EndFor
    \State $\hat{\mathbf{I}} \leftarrow \operatorname{TransformerBlocks}(\mathbf{Y}'')$ 
    \State \textbf{return} $\hat{\mathbf{I}}$  \hfill \textcolor{gray}{\# Restored document image}
    \end{algorithmic}
\end{algorithm}

\begin{algorithm}[!th]
    \caption{Pseudocode of DASAE}
    \label{alg:dasae_training}
    \begin{algorithmic}[1]
        
        \State \textcolor{gray}{\# Multi-branch disentanglement encoding}
        \For{$i = 1$ to $k$}
            \State $\mathbf{X}_{down, i}^0 \leftarrow \mathbf{I}$ 
            \For{$m = 1$ to $3$}
                \State $\mathbf{X}_i^m \leftarrow \operatorname{Conv}_{3 \times 3}(\operatorname{Conv}_{3 \times 3}(\mathbf{X}_{down, i}^{m-1}))$ 
                \State $\mathbf{X}_{down, i}^m \leftarrow \operatorname{MaxPool}(\mathbf{X}_i^m)$ \hfill \textcolor{gray}{\# Downsample}
            \EndFor
            \State $\mathbf{X}_i^4 \leftarrow \mathbf{X}_{down, i}^3$ \hfill \textcolor{gray}{\# Deepest feature representation}
        \EndFor
        
        \State \textcolor{gray}{\# Degradation perception gate (DPG)}
        \State $[w_1, \cdots, w_k] \leftarrow \operatorname{Softmax}(\mathcal{SE}(\mathbf{X}_1^4, \dots, \mathbf{X}_k^4))$
        
        \State \textcolor{gray}{\# Feature aggregation}
        \For{$m = 1$ to $3$} 
            \State $\mathbf{X}_{skip}^m \leftarrow \sum_{i=1}^k w_i \cdot \mathbf{X}_i^m$ \hfill \textcolor{gray}{\# Adaptive skip features}
        \EndFor
        \State $\mathbf{X}_{task} \leftarrow \sum_{i=1}^k w_i \cdot \mathbf{X}_i^4$ 
        \State $\mathbf{X}_{ASPP} \leftarrow \operatorname{ASPP}(\mathbf{X}_{task})$
        \State $\mathbf{X}_{in}^3 \leftarrow \mathbf{X}_{ASPP}$
        
        \State \textcolor{gray}{\# Structure decoding phase}
        \For{$m = 3$ to $1$}
            \State $\mathbf{X}_{up}^m \leftarrow \operatorname{ConvTranspose}(\mathbf{X}_{in}^m)$ \hfill \textcolor{gray}{\# Upsample}
            
            \State \textcolor{gray}{\# Feature fusion}
            \State $\mathbf{X}_{dec}^m \leftarrow 
            \operatorname{ConvBNReLU}(\operatorname{ConvBNReLU}(\left[\mathbf{X}_{up}^m, \mathbf{X}_{skip}^m\right]))$ 
            \If{$m > 1$}
                \State $\mathbf{X}_{in}^{m-1} \leftarrow \mathbf{X}_{dec}^m$
            \EndIf
        \EndFor
        
        \State $\mathbf{I}_M \leftarrow \operatorname{Sigmoid}(\operatorname{Conv}_{1 \times 1}(\mathbf{X}_{dec}^1))$
        
        \State \textbf{return} $\mathbf{I}_M$  \hfill \textcolor{gray}{\# Predicted structure map}
    \end{algorithmic}
\end{algorithm}

\begin{algorithm}[thpb] 
    \caption{Pseudocode of SWIM}
    \label{alg:swim_module}
    \begin{algorithmic}[1]
        \State \textcolor{gray}{\# Wavelet decomposition \& feature mapping}
        \State $(\mathbf{I}_{LL}, \mathbf{I}_{HL}, \mathbf{I}_{LH}, \mathbf{I}_{HH}) \leftarrow \mathcal{W}_{Haar}(\mathbf{I})$
        \State $(\mathbf{M}_{LL}, \mathbf{M}_{HL}, \mathbf{M}_{LH}, \mathbf{M}_{HH}) \leftarrow \mathcal{W}_{Haar}(\mathbf{I}_M)$
        \State $\mathbf{F}_L \leftarrow \operatorname{Conv}_{3\times3}(\mathbf{I}_{LL})$
        \State $\mathbf{F}_H^{rgb} \leftarrow \operatorname{Conv}_{3\times3}([\mathbf{I}_{HL}, \mathbf{I}_{LH}, \mathbf{I}_{HH}])$
        \State $\mathbf{F}_H^{st} \leftarrow \operatorname{Conv}_{3\times3}([\mathbf{M}_{HL}, \mathbf{M}_{LH}, \mathbf{M}_{HH}])$
        
        \State \textcolor{gray}{\# Frequency semantic aggregation}
        \State $\mathbf{F}'_L \leftarrow \operatorname{CA}(\mathbf{Q}\!=\!\mathbf{F}_L, \mathbf{K}\!=\!\mathbf{Y}, \mathbf{V}\!=\!\mathbf{Y})$
        \State $\mathbf{F}'^{rgb}_H \leftarrow \operatorname{CA}(\mathbf{Q}\!=\!\mathbf{F}_H^{rgb}, \mathbf{K}\!=\!\mathbf{Y}, \mathbf{V}\!=\!\mathbf{Y})$
        \State $\mathbf{F}'^{st}_H \leftarrow \operatorname{CA}(\mathbf{Q}\!=\!\mathbf{F}_H^{st}, \mathbf{K}\!=\!\mathbf{Y}, \mathbf{V}\!=\!\mathbf{Y})$
        
        \State \textcolor{gray}{\# Spatial gating to balance textures and structures}
        \State $\mathbf{W}_1, \mathbf{W}_2 \leftarrow \operatorname{G}([\mathbf{F}_H^{rgb}, \mathbf{F}_H^{st}])$ 
        \State $\mathbf{F}'_H \leftarrow \operatorname{Conv}_{1\times1}(\mathbf{W}_1 \odot \mathbf{F}'^{rgb}_H + \mathbf{W}_2 \odot \mathbf{F}'^{st}_H)$
        
        \State \textcolor{gray}{\# Begin of cross-frequency adaptive modulation (CFAM)}
       
        \State \quad \textcolor{gray}{\# Enhance high-frequency} 
        \State \quad $\mathbf{F}_H'' \leftarrow \operatorname{DWC}(\operatorname{DWC}(\operatorname{Conv}_{1\times1}(\mathbf{F}'_H)))$
        
        \State \quad \textcolor{gray}{\# Dual modulation} 
        \State \quad $\mathbf{F}''_L \leftarrow \operatorname{Conv}_{1\times1}(\mathbf{F}'_L)$
        \State \quad $\mathbf{G}_{map} \leftarrow \operatorname{Sigmoid}(\operatorname{Conv}_{1\times1}(\operatorname{DBD}(\mathbf{F}''_L)))$
        \State \quad $\boldsymbol{\gamma}, \boldsymbol{\beta} \leftarrow \operatorname{Conv}_{1\times1}(\operatorname{GAP}(\mathbf{F}''_L))$
        \State \quad $\tilde{\mathbf{F}}_H \leftarrow \boldsymbol{\gamma} \odot (\mathbf{F}_H'' \odot \mathbf{G}_{map}) + \boldsymbol{\beta}$
        \State \quad $\mathbf{F}_{out} \leftarrow \operatorname{Concatenation}(\tilde{\mathbf{F}}_H, \mathbf{F}'_L)$
        \State \textcolor{gray}{\# End of CFAM} 
        
        \State $\mathbf{Y} \leftarrow \operatorname{CA}(\mathbf{Q}\!=\!\mathbf{Y}, \mathbf{K}\!=\!\mathbf{F}_{out}, \mathbf{V}\!=\!\mathbf{F}_{out})$
        \State \textbf{return} $\mathbf{Y}$  \hfill \textcolor{gray}{\# High-frequency enhanced features}
    \end{algorithmic}
\end{algorithm}

Algorithm~\ref{alg:docpure_main} provides the pseudocode of our prompt-free unified document restoration framework (DocPure).
Algorithm~\ref{alg:dasae_training} outlines the pseudocode of our degradation-aware structure auto-encoder (DASAE). 
Algorithm \ref{alg:swim_module} describes the pseudocode of our structure-guided wavelet interaction module (SWIM).

\section{More experimental results}
\label{sec:comparisons}

\subsection{More ablation study}

\begin{table*}[htpb]
\centering
\caption{Ablation study of the internal components of DASAE on various document restoration tasks. The best performance is highlighted in \textbf{bold}.}
\label{tab:sp_ab}
\footnotesize
{
\begin{tabular}{c|ccc|cc|cc|cc|cc}
\hline
\multirow{2}{*}{{Configuration}} & 
\begin{tabular}[c]{@{}c@{}}Multi-branch\\ encoder\end{tabular} & 
\begin{tabular}[c]{@{}c@{}}Degradation\\ perception gate\\ (DPG)\end{tabular} & 
\begin{tabular}[c]{@{}c@{}}Routing\\ regularization\\ in DPG\end{tabular} & 
\multicolumn{2}{c|}{Deblurring} & 
\multicolumn{2}{c|}{Denoising} & 
\multicolumn{2}{c|}{CAR} & 
\multicolumn{2}{c}{Deshadowing} \\ 
\cline{5-12}
 & & & & PSNR$\uparrow$ & SSIM$\uparrow$ & PSNR$\uparrow$ & SSIM$\uparrow$ & PSNR$\uparrow$ & SSIM$\uparrow$ & PSNR$\uparrow$ & SSIM$\uparrow$ \\ 
\hline
Single encoder & \xmark & \xmark & \xmark & 31.98 & 0.9907 & 41.16 & 0.9972 & 31.17 & 0.9889 & 32.86 & 0.9525 \\
w/o DPG & \checkmark & \xmark & \xmark & 31.80 & 0.9904 & 41.18 & 0.9972 & 31.17 & 0.9888 & 32.83 & 0.9525 \\
w/o RR & \checkmark & \checkmark & \xmark & 32.41 & 0.9910 & 41.21 & 0.9972 & 31.30 & 0.9892 & 33.14 & 0.9527 \\ 
DASAE (ours) & \checkmark & \checkmark & \checkmark & \textbf{32.55} & \textbf{0.9912} & \textbf{41.26} & \textbf{0.9973} & \textbf{31.34} & \textbf{0.9893} & \textbf{33.31} & \textbf{0.9532} \\ 
\hline
\end{tabular}
}
\end{table*}

\begin{table*}[htbp]
\centering
\caption{Ablation study of CFAM on various document restoration tasks. The results demonstrate the performance gain across all four tasks when CFAM is employed. The best performance is highlighted in \textbf{bold}.}
\label{tab:ablation_wlfm}
\footnotesize
{
\begin{tabular}{c|cc|cc|cc|cc}
\hline
\multirow{2}{*}{\makebox[0.15\textwidth][c]{Configuration}} & 
\multicolumn{2}{c|}{\makebox[0.18\textwidth][c]{Deblurring}} & 
\multicolumn{2}{c|}{\makebox[0.18\textwidth][c]{Denoising}} & 
\multicolumn{2}{c|}{\makebox[0.18\textwidth][c]{CAR}} & 
\multicolumn{2}{c}{\makebox[0.18\textwidth][c]{Deshadowing}} \\ 
\cline{2-9} 
 & PSNR$\uparrow$ & SSIM$\uparrow$ & PSNR$\uparrow$ & SSIM$\uparrow$ & PSNR$\uparrow$ & SSIM$\uparrow$ & PSNR$\uparrow$ & SSIM$\uparrow$ \\ 
\hline
w/o CFAM & 31.46 & 0.9898 & 40.64 & 0.9967 & 30.59 & 0.9875 & 33.03 & 0.9528 \\
with CFAM (ours) & \textbf{32.55} & \textbf{0.9912} & \textbf{41.26} & \textbf{0.9973} & \textbf{31.34} & \textbf{0.9893} & \textbf{33.31} & \textbf{0.9532} \\ 
\hline
\end{tabular}
}
\end{table*}

\begin{figure}[!t]
    \centering
    \includegraphics[width=\columnwidth]{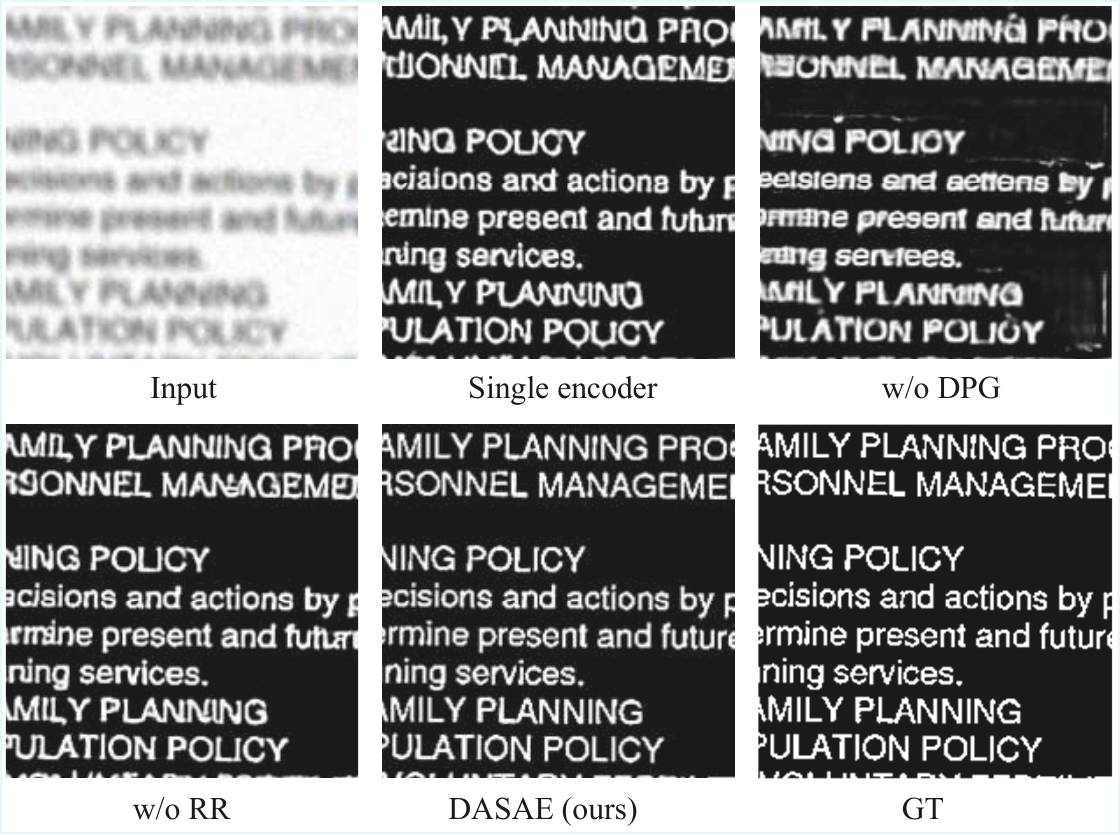}
    \caption{Visual comparison of ablation results for DASAE. The visualizations correspond to the ablation settings listed in Table~\ref{tab:sp_ab}. Utilizing a single encoder maps all degradation types into an entangled space. Furthermore, the absence of either DPG or routing regularization hinders the extraction of accurate degradation representations with blurry artifacts. In contrast, DASAE restores images with sharp structures.}
    \label{pic:ab_sp_encode}
\end{figure}

Table~\ref{tab:sp_ab} presents the ablation study of the internal components of DASAE\rev{, including the multi-branch encoder, the degradation perception gate (DPG), and the routing regularization (RR)}. The quantitative results demonstrate that, compared to the single encoder baseline or the variant without DPG employing simple average fusion, the proposed multi-branch encoder equipped with DPG without RR significantly improves the accuracy of  structure prediction. Furthermore, with the incorporation of routing regularization, the DPG can assign weights more precisely. This mechanism stabilizes the routing process, thereby further enhancing the efficacy of structure restoration.

Fig.~\ref{pic:ab_sp_encode} shows a visual comparison of ablation results for DASAE. The single encoder baseline forces the network to learn an entangled distribution of degradations. The blurred artifacts reveal the model's struggle in this entangled space. In contrast, the introduction of routing regularization in DASAE results in sharp structural maps, demonstrating that the degradation-discriminative routing strategy successfully prevents mode collapse and ensures task-specific feature extraction.

Table~\ref{tab:ablation_wlfm} illustrates the performance gains attributed to CFAM. The enhancement is particularly pronounced in the deblurring task, which requires greater fidelity in restoring high-frequency details. This indicates that CFAM plays a crucial role in preserving structural textures. 
CFAM adaptively modulates the response of high-frequency features via spatial gating and channel modulation. This cross-frequency interaction ensures that the restored high-frequency details are consistently aligned with the low-frequency content in terms of global structure.

\subsection{More visual results}
\begin{figure*}[!th] 
    \centering
    \includegraphics[width=\textwidth]{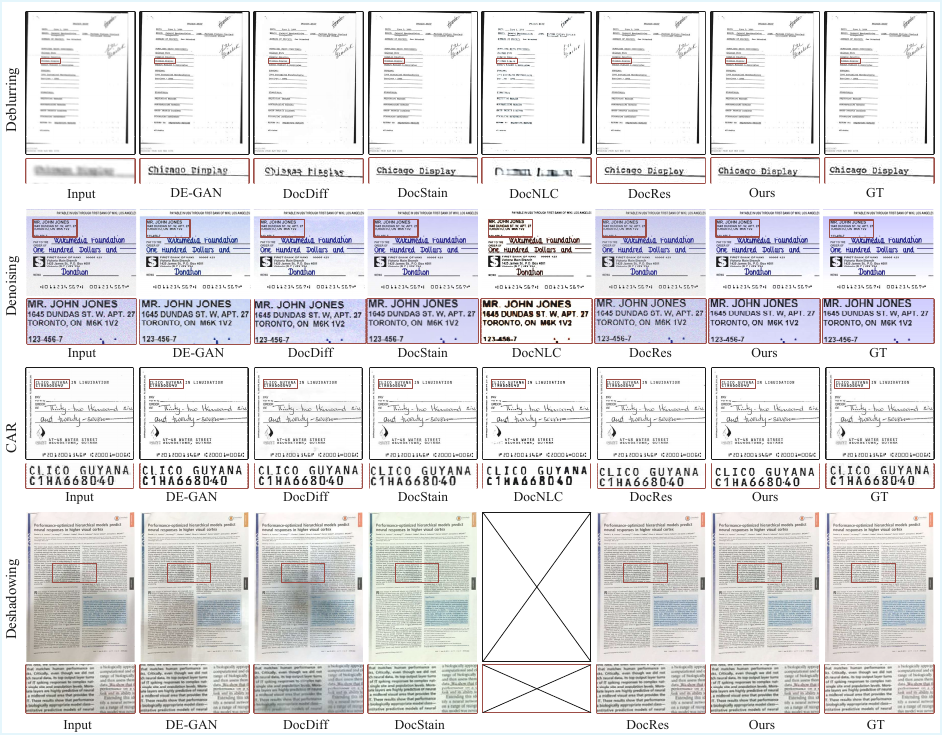}
    \caption{More qualitative comparison across four restoration tasks: deblurring, denoising, compression artifact reduction (CAR), and deshadowing. The results are reported on the FUNSD$_{Blur}$, BCSD$_{Noise}$, BCSD$_{CAR}$, and Jung's dataset~\cite{jung2018jun}, respectively. Note that these datasets feature document styles distinct from the training set. {Benefiting from the structure predicted by DASAE and the frequency-domain interaction facilitated by SWIM, our method achieves higher text structural fidelity and better background consistency.}}
    \label{fig:compare_car}
\end{figure*}

As shown in Fig.~\ref{fig:compare_car}, to further demonstrate the robustness of our method, we provide additional qualitative comparisons against the state-of-the-art (SOTA) methods on four primary document restoration tasks: deblurring, denoising,  compression artifact reduction (CAR), and deshadowing. The compared SOTA methods include general-task restoration models (DE-GAN~\cite{souibgui2020gan}, DocDiff~\cite{yang2023docdiff}, and DocStain~\cite{li2025high}), and all-in-one restoration models (DocNLC~\cite{wang2024docnlc} and DocRes~\cite{zhang2024docres}). {For the deshadowing task, since DocNLC requires cross-task pairing for the training data, it cannot be trained with the unpaired deshadowing dataset.}

\begin{figure*}[!ht]
    \centering
    \includegraphics[width=\textwidth]{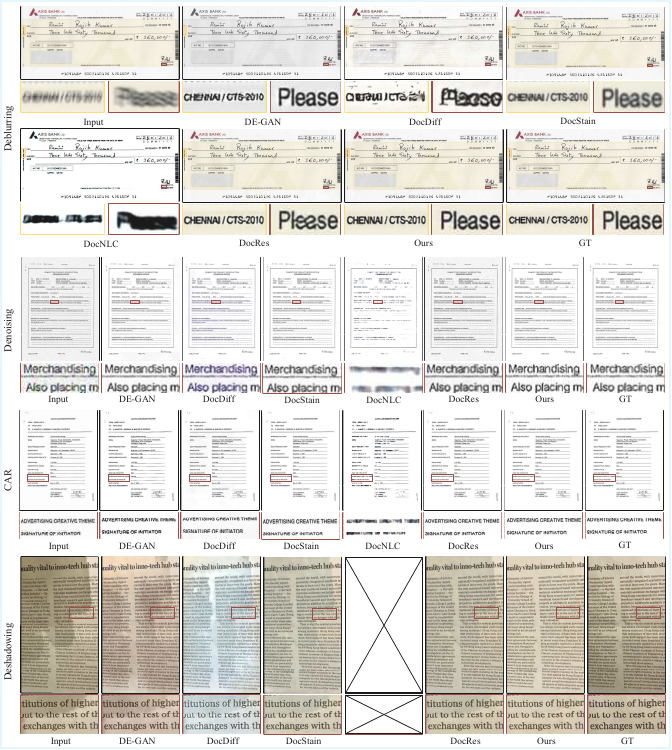}
    \caption{Qualitative comparison of generalization capabilities across four image restoration tasks: deblurring, denoising, compression artifact reduction (CAR), and deshadowing, evaluated on the BCSD$_{blur}$, FUNSD$_{noise}$, FUNSD$_{CAR}$, and OSR datasets, respectively. The evaluation is conducted on document images with styles distinct from the training set. {Leveraging our degradation-aware structure-guided wavelet modulation, our DocPure exhibits enhanced generalization to text patterns and complex backgrounds.}}
    \label{fig:compare_add}
\end{figure*}

Fig.~\ref{fig:compare_add} illustrates the robustness of our model against distribution shifts. 
In contrast to existing SOTA methods, which are prone to structural distortions or residual artifacts in cross-domain scenarios, DocPure exhibits superior generalization capabilities, enabling the high-quality restoration of clear text content.

\bibliographystyle{unsrt}

\bibliography{docpure}

\end{document}